\documentclass[10pt,twocolumn,letterpaper]{article}

\usepackage[pagenumbers]{cvpr} % To force page numbers, e.g. for an arXiv version

\definecolor{cvprblue}{rgb}{0.21,0.49,0.74}
\usepackage[pagebackref,breaklinks,colorlinks,allcolors=cvprblue]{hyperref}

\usepackage[linesnumbered, ruled]{algorithm2e}
\SetKwRepeat{Do}{do}{while}

\def\paperID{16101}
\def\confName{CVPR}
\def\confYear{2025}

\title{Hierarchical Compact Clustering Attention (COCA) for Unsupervised Object-Centric Learning}

\author{
    Can Küçüksözen$^{1,2}$ \quad Yücel Yemez$^{1,2}$ \\
    $^1$Department of Computer Engineering, Koç University \\
    $^2$KUIS AI Center\\
    {\tt\small \{ckucuksozen19, yyemez\}@ku.edu.tr}
}

\begin{document}
%\maketitle

\twocolumn[{%
\renewcommand\twocolumn[1][]{#1}%
\maketitle
\begin{center}
    \centering
    \captionsetup{type=figure}
    \vspace*{-1.em}
    \fbox{\includegraphics[width=1.\textwidth,height=.25\textwidth]{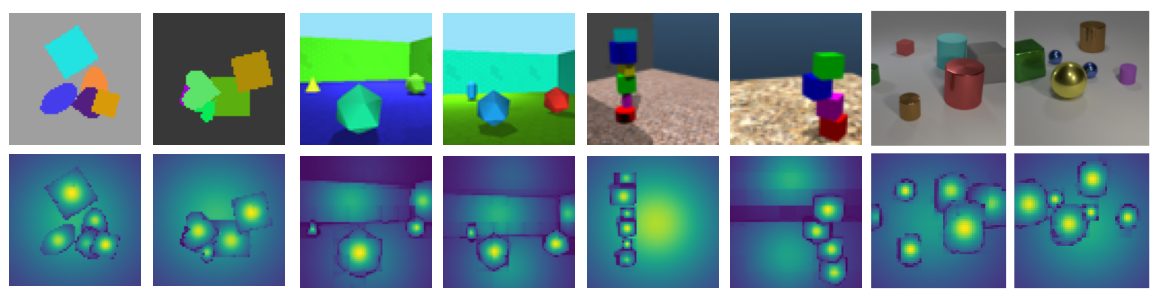}}
    \captionof{figure}{Compactness scores obtained for each pixel in the scene, across four different datasets. The transition from bright yellow to deep purple signifies decreasing compactness. To obtain these scores, a trained COCA-Net encoder is used to generate object masks. Each object mask is then broadcasted to pixels based on the pixel-object assignments. This operation associates every pixel with a copy of its object's mask. Finally, compactness scores for each pixel's mask are calculated via Eq. \ref{eq:compactness_of_affinities}.}
    \label{fig:comp_vis}
\end{center}%
}]
\vspace*{-1.em}
\begin{abstract}
 We propose the Compact Clustering Attention (COCA) layer, an effective building block that introduces a hierarchical strategy for object-centric representation learning, while solving the unsupervised object discovery task on single images. COCA is an attention-based clustering module capable of extracting object-centric representations from multi-object scenes, when cascaded into a bottom-up hierarchical network architecture, referred to as COCA-Net. At its core, COCA utilizes a novel clustering algorithm that leverages the physical concept of compactness, to highlight distinct object centroids in a scene, providing a spatial inductive bias. Thanks to this strategy, COCA-Net generates high-quality segmentation masks on both the decoder side and, notably, the encoder side of its pipeline. Additionally, COCA-Net is not bound by a predetermined number of object masks that it generates and handles the segmentation of background elements better than its competitors. We demonstrate COCA-Net's segmentation performance on six widely adopted datasets, achieving superior or competitive results against the state-of-the-art models across nine different evaluation metrics.
\end{abstract}    
\section{Introduction}
\label{sec:intro}

Object-centric learning (OCL) has emerged as a powerful paradigm in contemporary computer vision, offering advantages such as improved generalization over supervised methods, robustness to data distribution shifts, effectiveness in downstream tasks, and the ability to train without labeled data \cite{Dittadietal22,Greff2020OnTB}. These models allocate a \textit{separate representational slot} for each object in a scene, enabling the disentanglement of individual object properties in multi-object scenarios. The benefits of OCL have driven its increasing adoption across diverse applications. While these models achieve impressive object discovery accuracy on synthetic datasets, their performance—and consequently their generalization capabilities—deteriorates significantly when applied to real-world images, which are far more complex and diverse. Yang et al., \cite{benchmarking_Yang2024} argue that these models still lack the necessary inductive biases to understand ‘objectness’ in real-world images.

In fact, the challenges of contemporary OCL models extend beyond real-world images, with critical limitations also emerging in synthetic datasets. State-of-the-art models \cite{sa, ShepherdingKim2023, impsa, boqsa} derived from Slot-Attention (SA) \cite{sa} have gained popularity for their efficiency and simplicity. However, recent studies highlight key drawbacks, such as learning inconsistencies due to similar initialization of slot representations from a common latent distribution \cite{ShepherdingKim2023, unlockingZhang23}, sensitivity to a predefined number of slots \cite{fan2024adaptive, SensitivityOf2023} and poor handling of background segments. We argue that these issues stem from how \textit{segregation}—the process of organizing raw sensory inputs into distinct, meaningful entities—is handled. When applied to our case, segregation corresponds to unsupervised instance segmentation, where the model partitions pixels into coherent object segments. Most often, unsupervised segmentation is addressed through the rich literature on cluster analysis, where SA can be seen as a \textit{soft} version of the traditional K-Means clustering algorithm \cite{sa}. Thus,  inheriting its strengths and limitations, including sensitivity to initialization, dependency on a predefined number of clusters, assumptions about cluster shape and size, and sensitivity to outliers  \cite{everitt2011cluster,Jain1988AlgorithmsFC,Xu2005SurveyOC,hastie2009elements,kaufman2009finding}, which hinder SA's suitability for complex or heterogeneous scenes.

In this work, we address the limitations of the current unsupervised OCL paradigm by developing a novel approach that avoids the aforementioned drawbacks of SA based methods. We define a notion of ‘objectness’, as a spatial inductive bias based on \textit{compactness of candidate object masks} and incorporate it into our architecture. We lay the foundations of our proposed architecture on the hierarchical agglomerative clustering (HAC) based methods—which have not received much attention from the research community in recent years— despite its inherent advantages that include flexibility in the number of output clusters, robustness to noise and outliers, the ability to capture hierarchical data relationships, smooth handling of irregularly distributed clusters, and high interpretability.
\cite{ComprehensiveSurveyHAC,everitt2011cluster,Xu2005SurveyOC,hastie2009elements}. 

To this end, we introduce the Compact Clustering Attention Layer (COCA layer), a simple and effective attention-based clustering module. When stacked hierarchically, the COCA layers form the COCA-Net image encoder. Combined with the Spatial Broadcast Decoder (SBD) \cite{Watters2019SpatialBD}, COCA-Net adopts a bottom-up approach for end-to-end unsupervised object discovery from scratch.
\vspace{-7pt}
\paragraph{Contributions} To the best of our knowledge, in the context of unsupervised OCL literature, COCA-Net is the first work to utilize a hierarchical clustering and pooling strategy within a neural network architecture to generate slot representations for a single image, in a fully unsupervised way. Due to its design, COCA-Net offers several key advantages, addressing most of the drawbacks that existing slot-based methods suffer from:
\begin{itemize}
\item Robustness to initial seeding: Different training runs of the COCA-Net hierarchy yield almost the same performance with minimal variance, demonstrating its reliability and robustness. 
\item Effective background segmentation: COCA-Net effectively handles background segments, ensuring accurate segmentation of the entire scene.
\item  Dynamic slot allocation: Each COCA layer can dynamically adjust the number of its output clusters, eliminating the need to predetermine the number of output object slots.
\item High-quality encoder features: COCA-Net generates high-quality segmentation masks on the encoder side. This holds promise for the trained COCA-Net encoder to be repurposed as an object-centric feature extractor for downstream tasks.
\end{itemize}

We conduct extensive experiments on the OCL benchmark provided by \cite{Dittadietal22} and demonstrate that COCA-Net outperforms or performs on par with the state-of-the-art on the unsupervised object discovery task. This evaluation is based on the object slot masks obtained from both the decoder and encoder sub-networks of the pipeline, in contrast to literature on OCL where the evaluation considers only the masks generated by the decoder sub-network.

\section{Related Work}
\label{sec:related}
\subsection{Unsupervised Object Discovery}

Recent years have seen a surge in \textit{Instance Slot} models, notably Slot Attention (SA) \cite{sa} and its successors \cite{SpotlightAR, unlockingZhang23, ShepherdingKim2023, fan2024adaptive, dinasaur, BMOD, boqsa, impsa, invsa}. These models typically initialize a fixed number of slots from a shared latent distribution and use query-normalized cross-attention to assign input features to slots iteratively, refining slot features in the process. 

However, initializing slots from a common distribution causes what is known as the \textit{routing problem}, where symmetry among similarly initialized slots must be broken to separate distinct objects \cite{Greff2020OnTB}. Zhang et al. \cite{unlockingZhang23} highlight that the routing problem -- or the lack of tie-breaking in Slot Attention (SA) methods -- can cause multiple slots to bind to the same input subset, hindering object separation. Similarly, Kim et al. \cite{ShepherdingKim2023} describe the \textit{bleeding issue}, where different training runs yield inconsistent results due to SA’s occasional failure to distinguish foreground objects from the background. Beyond challenges related to the routing problem, a key limitation of SA is its requirement to predetermine the number of output slots. As noted in recent studies \cite{SensitivityOf2023,fan2024adaptive}, predefining the number of objects in a dataset is often impractical since scenes can contain a variable number of objects. Consequently, performance of SA-based models is highly sensitive to the number of slots that are predefined; poor choices can lead to under- or over-segmentation. These limitations in SA methods arise from their reliance on a soft K-Means based clustering approach, which shares the well-documented drawbacks of traditional K-Means \cite{everitt2011cluster,Jain1988AlgorithmsFC,Xu2005SurveyOC,hastie2009elements,kaufman2009finding, DigitalImageProcessing}.

\textit{Sequential Slot} models like \cite{MONetBurgess2019,GENESIS_Engelcke2019,GENESISV2_Engelcke2021}, on the other hand, are specifically tailored to overcome the routing problem by imposing a sequential masking -- or concealing -- strategy to discover distinct object masks. In each iteration, a new object mask is predicted, and the explained parts of the scene are excluded from subsequent predictions via Stick-Breaking Clustering (SBC) \cite{GENESISV2_Engelcke2021}. To generate a single object mask, for instance, \cite{GENESISV2_Engelcke2021} samples a random pixel in the scene and computes its affinity mask whereas \cite{MONetBurgess2019} relies on a randomly initialized sub-network. Although effective to mitigate the routing problem, these models suffer from suboptimal mask generation and high computational costs per object, limiting their performance in complex object-rich scenes. The third category of slot based methods, \textit{Spatial Slots}, partition the scene into spatial windows and assign a distinct slot to each window. By focusing each slot on a distinct local area, these models effectively break slot symmetries and mitigate the routing problem. Notable examples include \cite{SPACE_Lin2020, SPAIR_Crawford2019, SCALOR_Jiang2019}. However, their performance deteriorates when dealing with overlapping objects or objects larger than the predefined slot windows \cite{Greff2020OnTB}.

\subsection{Clustering Methods}

Our work is based upon the traditional literature on \textit{hierarchical clustering} methods \cite{Arbelaez_ContourDetectHierarchicalImgSeg, Uijlings_2013}, which iteratively cluster pixels across multiple scales, often using a bottom-up agglomerative approach. Current hierarchical agglomerative clustering methods based on deep learning primarily focus on segmentation in real-world image datasets, typically optimizing weakly supervised or self-supervised objectives \cite{hassod_cao2023, sohes_cao2024, CAST_recforseg_ke2024, UnSAMwang2024}. Most of them make use of pre-trained backbones or sub-networks \cite{sohes_cao2024, hassod_cao2023, UnSAMwang2024} or teacher networks to guide learning \cite{sohes_cao2024, hassod_cao2023}. Differing from these models is a recent method, \cite{HSG_multivewcoseg}, which leverages multi-views of a scene to provide additional supervision. 

A relevant part of the literature is the methods that operate on graph structured data and address \textit{graph-based clustering} and pooling. Some recent approaches can be listed as \cite{Bianchi2019SpectralCW, visionGNN_han2022, DMON_Tsitsulin2023, TokenCutWang2023, CutLerWang2023}. Notably, TokenCut \cite{TokenCutWang2023}, uses a pre-trained backbone \cite{DINO_Caron2021} to initialize node features and applies a Normalized Cut \cite{shiandmalik} based spectral graph clustering algorithm to achieve unsupervised object discovery in scenes with a single prominent object. \cite{CutLerWang2023} and \cite{Cuvler} adopt TokenCut to multi-object scenes by leveraging an SBC inspired concealing strategy. A hybrid hierarchical and graph based clustering model \cite{UnSAMwang2024}, uses TokenCut \cite{CutLerWang2023} to generate a top-to-bottom partition of the scene and utilizes a form of \cite{sohes_cao2024} as a hierarchical agglomerative clustering to produce bottom-to-top refining of the initial partitions.

\begin{figure*}
  \centering
  \begin{subfigure}{0.70\linewidth}
  \centering
    \fbox{\includegraphics[width=1.\textwidth,height=.6\textwidth]{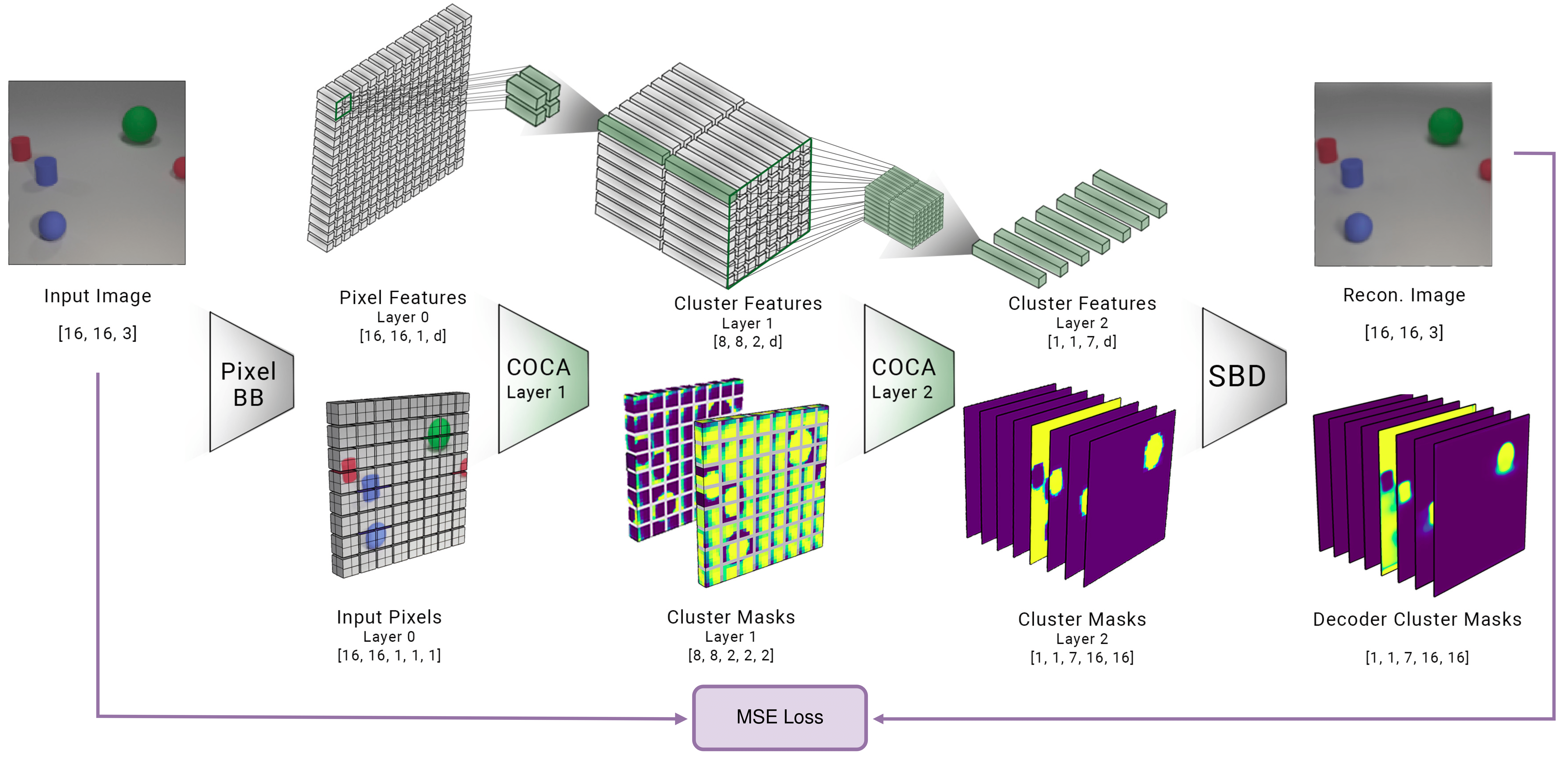}}
    \caption{Overview of the COCA-Net Hierarchy: COCA-Net processes a $16 \times 16$ input image and corresponding features from the pixel encoder backbone. The first COCA layer partitions these features into $2 \times 2$ non-overlapping windows, performing clustering and pooling in parallel to produce $8 \times 8 \times 2 \times d$ cluster features. Layer 1 outputs two clusters, with $2 \times 2$ cluster masks aggregated to the next layer along with pooled features. Similarly, the second COCA layer partitions its input into an $8\times8$ window and outputs 7 clusters. Finally, the image is reconstructed by the SBD, where the loss is Mean Squared Error.}
    \label{fig:short-a}
  \end{subfigure}
  \hfill
  \begin{subfigure}{0.27\linewidth}
    \centering\fbox{\includegraphics[width=1.\textwidth,height=1.4\textwidth]{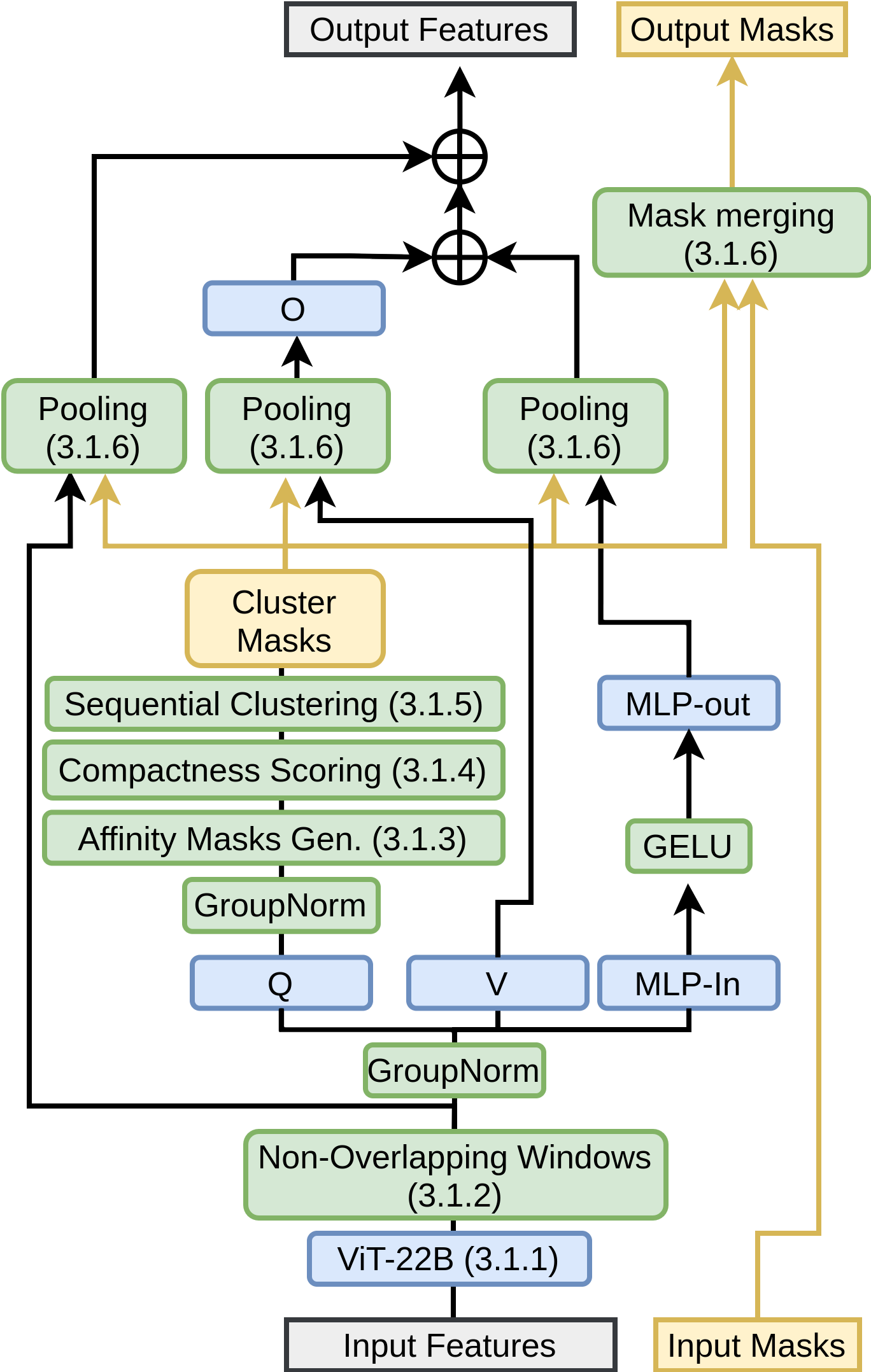}}
    \caption{Best viewed in color. Overview of a single COCA layer, which can be seen as a clustering variant of the ViT-22B layer \cite{vit22b_dehghani2023}. Blue boxes indicate layers with learnable parameters, green boxes denote non-parametric operations, and yellow boxes represent operations involving masks. }
    \label{fig:short-b}
  \end{subfigure}
  \caption{Overall summary of the COCA-Net hierarchy and a single COCA layer.}
  \label{fig:short}
\end{figure*}

\section{Method}
\label{sec:method_3_0}

One of the key insights behind the COCA layer aligns with the principles of \textit{perceptual grouping} in vision systems \cite{shiandmalik, Arbelaez_2014, Uijlings_2013, deng2024perceptual}, where an ideal unsupervised segmentation model (e.g., COCA-Net) is expected to produce similar feature representations for pixels within the same object, while distinct features emerge for pixels from different objects or the background.
To be able to learn such a feature space, we employ a simple pixel feature encoder as our backbone. Recent findings support this point-wise handling of pixels over locality-based inductive biases, \cite{16by16patches_Nguyen2024} reports that the local patchification process used in the state-of-the-art Vision Transformer architectures \cite{ViT_Dosovitskiy2021} is, in fact, unnecessary. Thus, our backbone initializes pixel features that encode both appearance and position information. The details of this pixel feature encoder can be found in Supplementary Material.

\subsection{COCA Layer}
\label{subsec:coca}
If a model achieves the ideal scenario described by the perceptual grouping literature, the task of unsupervised object discovery can be simplified to identifying a single \textit{anchor -- or representative -- node} per object, whose \textit{affinity mask} would then represent the object's segmentation. This identification of an anchor node per object could potentially benefit from insights into what constitutes a visual object and their compositional nature in the physical world. Building on such intuitions, a single COCA layer embodies a five-step approach to hierarchical agglomerative clustering, also illustrated in Figure \ref{fig:short-b}; 
\begin{enumerate}
    \item Refining the features of each input node within its spatial neighborhood to relate and contrast appropriate elements, 
    \item Partitioning these refined nodes into non-overlapping windows,
    \item Generating a set of \textit{affinity masks} within each window in parallel,
    \item Evaluating these affinity—candidate object—masks based on their \textit{compactness scores},
    \item Running an SBC-based sequential concealing strategy to iteratively identify the most compact mask, assign its constituent nodes as the next output cluster, and conceal these already clustered nodes to prevent potential duplicate masks in subsequent iterations.
\end{enumerate}
The layer is finalized by pooling the features of nodes that correspond to distinct output clusters, merging masks that are produced in the last two consecutive layers and aggregation of these elements to the next level of the hierarchy.  

\subsubsection{Feature Refinement} 
\label{sec:feature_refinement}
For effective and adaptive feature refinement, COCA leverages the dynamic message-passing capabilities of self-attention \cite{Attention_Vaswani2017} layers in Vision Transformers (ViTs) \cite{ViT_Dosovitskiy2021}. The widespread success of ViTs across various tasks and modalities makes them an ideal choice for refining features among neighboring nodes within the same object. Specifically, we leverage the self-attention operation that is used in a recent version, called  ViT-22B \cite{vit22b_dehghani2023}, which modifies and scales the original model to 22 billion parameters. This refinement process enhances feature similarity between nodes that are likely to be clustered together, first locally at earlier hierarchy levels and then globally at higher levels with an expanded receptive field, allowing spatial patterns in the data that correspond to distinct objects to be gradually distinguished. Implementation details of the feature refinement stage can be found in Supplementary Material.

\subsubsection{Operating on Non-Overlapping Windows}
\label{sec:non_overlapping_partitioning}
At each layer $l$, where $l=1,2,...,L$, COCA handles a total of  $h_{l-1}^\prime \times w_{l-1}^\prime \times k_{l-1}$ nodes refined by the vision transformer, where $h_{l-1}^\prime$, $w_{l-1}^\prime$ are input spatial dimensions and $k_{l-1}$ is input cluster dimension. 
%The input cluster dimension for layer $l=1$ is simply equal to one, i.e., $k_0 = 1$. 
COCA partitions these nodes into $t^2_l$ non-overlapping windows, where each window is composed of $h_l\times w_l \times k_{l-1}$ nodes. Note that $h_l=h_{l-1}^\prime/t_l$ and $w_l=w_{l-1}^\prime/t_l$. This partitioning results in an unfolded feature tensor for each window $t$ at layer $l$, which we represent by $\mathbf{X}^{l,t} \in \mathbb{R}^{n_l \times d_l}$ where $n_l = h_l \times w_l \times k_{l-1}$ and $d_l$ represents the feature dimension of a node. In effect, COCA maps $n_l = h_l \times w_l \times k_{l-1}$ input nodes to $1 \times 1 \times k_l$ distinct output clusters, for each window, in parallel. The clusters generated at the highest layer $L$ yield the final object slots. Note that we henceforth omit the window index $t$ for ease of notation since each window is treated simultaneously in the same manner.

\subsubsection{Generating Candidate Affinity Masks} 
\label{sec:affinity_masks}
The \textit{affinity masks} in a COCA layer, denoted by $\mathbf{\Lambda}^{l} \in \mathbb{R}^{n_l \times n_l}$, represent the pairwise affinities between nodes $i$ and $j$ within a window, where $i=1,2,...,n_l$ and $j=1,2,...,n_l$. Each entry $\mathrm{\Lambda}^{l}_{ij}$ in this affinity mask tensor is a soft binary value between 0 and 1, indicating how similar nodes $i$ and $j$ are. To compute pairwise affinities between each node pair, we begin by projecting the refined input node features onto a common high-dimensional space and apply GroupNorm normalization \cite{GroupNorm}. Next, we calculate the Euclidean distance $\mathrm{E}^{l}_{ij}$ between nodes $i$ and $j$ based on their feature representations after projection and normalization:
\begin{equation}
\label{eq:euc_distances}
\mathrm{E}^{l}_{ij} =  \frac{\tau_l}{\sqrt{n_l \cdot d_l}} \cdot \Big\| \mathbf{Y}^{l}_i - \mathbf{Y}^{l}_j \Big\|^2 
\end{equation}
where $\tau_l$ is the temperature hyper-parameter for layer $l$ and $\mathbf{Y}^{l} \in \mathbb{R}^{n_l \times d_l}$ denotes the projected and normalized feature vectors of nodes. Each column vector $\mathbf{E}^{l}_i \in \mathbb{R}^{n_l}$ in $\mathbf{E}^{l}$ can be viewed as a distance mask of a single node $i$ measured against all other nodes within the window. So we apply soft-argmin normalization followed by min-max scaling to each distance mask, obtaining the affinity mask $\mathbf{\Lambda}^{l}_i \in \mathbb{R}^{n_l}$ for each node $i$ (see Supplementary Material for details). This normalization procedure dynamically builds associations between nodes within a window and maps each affinity value into the required range $[0,1]$ for compactness scoring and sequential clustering, which we describe next.

\subsubsection{Compactness Scoring}
\label{sec:compactness}
Compactness can be viewed as a scoring function that operates on affinity masks. A single affinity mask has dimensions equal to the number of input nodes within a window, inherently forming a two-dimensional shape. Compactness scoring function is intended to guide the model to prioritize nodes that produce affinity masks corresponding to compact shapes, assigning them higher scores. Intuitively, a prominent foreground object is expected to have a compact and convex spatial form, while background elements are often scattered, with holes and concave edges. Various approaches exist across diverse disciplines for measuring compactness. A notable approach \cite{MassamGoodchild1971,EfficientMesCompactness_Li2013, MassCompactness} for calculating the \textit{mass normalized compactness} of a shape -- or affinity mask -- $\boldsymbol{\Lambda}$, is based on the concept of moment of inertia (MI), which physically quantifies shape dispersion relative to a point, in a scale-invariant and additive manner. The compactness functional $\mathcal{C}^{\mu}(\boldsymbol{\Lambda})$ of $\boldsymbol{\Lambda}$ around point $\mu$ can be computed by comparing its MI to that of the most compact two-dimensional shape in this scenario: a circle $\mathbf{\Theta}$ with the same effective area as $\boldsymbol{\Lambda}$ and centered at $\mu$:
\begin{equation}
\label{eq:compactness}
\mathcal{C}^{\mu}(\boldsymbol{\Lambda}) = \frac{\mathcal{I}^{\mu}(\mathbf{\Theta}_\mu)}{\mathcal{I}^{\mu}(\boldsymbol{\Lambda})}
\end{equation}
where $\mathcal{I}^{\mu}(\mathbf{\Theta}_\mu)$ represents the MI computed around $\mu$, for the circle $\mathbf{\Theta}_\mu$ that is centered at $\mu$. This compactness measure is bounded within $(0,1]$, with larger values indicating more compact shapes. A perfect circle achieves the maximum compactness value of 1.

Formally, we treat the affinity masks tensor $\mathbf{\Lambda}^{l} \in \mathbb{R}^{n_l \times n_l}$ as a collection of singular affinity masks $\boldsymbol{\Lambda}^l_i \in \mathbb{R}^{n_l}$, each associated with a node $i$. A single affinity mask can be considered as a flattened two dimensional grid of affinities that constitute a non-uniform density shape. To simulate the physical properties of nodes that construct the affinity masks, during the execution of the first COCA layer, we declare five attributes for each pixel; area $\mathbf{A}^{1} \in \mathbb{R}^{n_1\times1}$, mass $\mathbf{M}^{1} \in \mathbb{R}^{n_1\times1}$, density $\mathbf{D}^{1} \in \mathbb{R}^{n_1\times1}$, moment of inertia $\mathbf{I}^{1} \in \mathbb{R}^{n_1\times1}$ and mean position in the original image resolution $\mathbf{P}^{1} \in \mathbb{R}^{n_1 \times 2}$. Note that $D^l_i = M^l_i/A^l_i$. After the clustering is completed, these attributes are pooled according to the cluster masks and aggregated to the next levels of the hierarchy, just as the cluster feature vectors. 

In order to distribute a copy of these physical attributes of nodes to each affinity mask and scale these attributes based on the corresponding affinities, we compute the intermediate attributes, $\tilde{\mathbf{A}}^{l}, \tilde{\mathbf{D}}^{l}, \tilde{\mathbf{I}}^{l}, \tilde{\mathbf{M}}^{l} \in \mathbb{R}^{n_l \times n_l}$, by first broadcasting then computing an element-wise product with the affinity masks $\mathbf{\Lambda}^{l} \in \mathbb{R}^{n_l \times n_l}$. The details for this broadcasting and scaling operations are shared in Supplementary Material. 
With the intermediate attributes of each affinity mask are computed, inspired by \cite{MassamGoodchild1971,EfficientMesCompactness_Li2013, MassCompactness}, we compute the \textit{mass normalized compactness}, $\mathcal{C}^{i}(\boldsymbol{\Lambda}^l_i)$, of the shape described by the affinity mask $\boldsymbol{\Lambda}^l_i \in \mathbb{R}^{n_l}$ around the axis passing through node $i$ as:
\begin{equation}
\label{eq:compactness_of_affinities}
\mathcal{C}^{i}(\boldsymbol{\Lambda}^l_i) = \frac{
\sum_{j} \tilde{\mathrm{M}}^{l}_{ij}\tilde{\mathrm{A}}^{l}_{ij} + \sum_{j<v} 2\min(\tilde{\mathrm{D}}^{l}_{ij},\tilde{\mathrm{D}}^{l}_{iv}) \tilde{\mathrm{A}}^{l}_{ij}\tilde{\mathrm{A}}^{l}_{iv}}
{2 \pi \cdot \left( \sum_{j} \tilde{\mathrm{I}}^{l}_{ij} + \tilde{\mathrm{M}}^{l}_{ij} \Delta^{l}_{ij} \right)}
\end{equation}
where $\mathbf{\Delta}^{l}_{i} \in \mathbb{R}^{n_l}$ represents the Euclidean distances from the position of node $i$ to all other nodes, with entries $\Delta^{l}_{ij}$ computed as:
\begin{equation}
\Delta^{l}_{ij} = \big\| \mathbf{P}^{l}_{i} - \mathbf{P}^{l}_{j} \big\|_2^2 \
\end{equation}
The compactness measure derived from the moment of inertia offers several advantages for our hierarchical clustering architecture. First, it is additive, meaning that the compactness of a complex shape can be efficiently computed as a linear combination of the compactness values of its parts. Additionally, as noted in \cite{EfficientMesCompactness_Li2013}, this measure is insensitive to shape size and complex boundaries. 
By assigning higher scores to compact shapes, the model is encouraged to prioritize regions corresponding to well-defined foreground objects. It is important to note that we use Equation \ref{eq:compactness_of_affinities} to compute compactness for all affinity masks across all windows in layer $l$. These computations are performed in parallel and only once per layer, prior to the sequential cluster generation process detailed next. This strategy provides an efficiency advantage over existing Sequential Slot models or recent graph based clustering methods, possibly much more when dealing with crowded scenes.

\subsubsection{Sequentially Discovering Object Centroids} \label{sec:seq_clus} A key feature of MI-based compactness measurement emerges when considering the node around which the MI is calculated. When this node coincides with the \textit{centroid} $\gamma$ of the shape $\boldsymbol{\Lambda}$, the MI value $I^\gamma(\boldsymbol{\Lambda})$ reaches its minimum, which translates to maximum compactness achievable for any node within $\boldsymbol{\Lambda}$ \cite{EfficientMesCompactness_Li2013}. 
This property of MI-based compactness offers a significant advantage for unsupervised scene segmentation. Nodes corresponding to the centroids of distinct objects will yield the highest compactness scores within their respective objects, as illustrated with Figure \ref{fig:comp_vis}. By leveraging this spatial inductive bias introduced by compactness measurement, the clustering algorithm can effectively focus on the centroids of distinct objects, facilitating object separation even in challenging cases where closely located objects have similar appearances.

Specifically, our cluster generation algorithm can be considered as a variation of SBC implemented in \cite{MONetBurgess2019,GENESISV2_Engelcke2021}. Motivated by these works, we start by initializing a \textit{scope} tensor $\mathbf{Z}^l_0 \in \mathbb{R}^{n_l}$ at iteration $m=0$ as a tensor full of ones with the same dimensions as the compactness scores. Following \cite{MONetBurgess2019, GENESISV2_Engelcke2021}, the scope tensor $\mathbf{Z}^l_m$  represents a map of input nodes which are still unassigned and awaiting clustering in iteration $m$ of the sequential clustering process.  Given $\mathbf{C}_0^l \in \mathbb{R}^{n_l}$ and $\mathbf{Z}^l_0 \in \mathbb{R}^{n_l}$, the algorithm proceeds as follows at each iteration $m$: i) The compactness scores are eroded by element-wise multiplication with the current scope tensor $\mathbf{Z}^l_m$, ensuring that only unassigned nodes (with non-zero scope values) contribute to subsequent calculations, (Eq. \ref{eqn:line-1}). ii) The node $\omega_m$ that generates the affinity mask $\mathbf{\Lambda}^l_{\omega_m}$ with the highest compactness score is selected as the anchor of the current cluster, (Eq. \ref{eqn:line-2}). iii) The affinity mask $\mathbf{\Lambda}^l_{\omega_m}$ associated with the selected anchor node is concealed by the current scope and designated as the next output cluster $\mathbf{\Pi}^l_{m} \in \mathbb{R}^{n_l}$, (Eq. \ref{eqn:line-3}). iv) The scope tensor $\mathbf{Z}^l_{m}$ is updated by masking out nodes included in $\mathbf{\Pi}^l_{m}$, excluding these nodes from further iterations, (Eq. \ref{eqn:line-4}). One iteration of the cluster generation process can be summarized as follows (omitting the layer index $l$):
\begin{subequations}
\begin{align}
    \mathbf{C}_{m} &= \mathbf{C}_{(m-1)} \odot \mathbf{Z}_{(m-1)} \label{eqn:line-1} \\
    \omega_m &= \mathrm{argmax}_i \left( \mathrm{C}_{mi} \right) \label{eqn:line-2}\\
    \mathbf{\Pi}_{m} &= \mathbf{\Lambda}_{\omega_m}  \odot  \mathbf{Z}_{m-1} \label{eqn:line-3}\\
    \mathbf{Z}_{m} &= \mathbf{Z}_{(m-1)} \odot \left(\ \mathbf{1}^{n_l} - \mathbf{\Pi}_{m} \right) \label{eqn:line-4}
\end{align} 
\label{eqn:sequential}
\end{subequations}

\vspace{-10pt}
\subsubsection{Pool, Aggregate and Skip Connect}
\label{sec:pool}
Following the generation of output cluster masks $\mathbf{\Pi}^{l} \in \mathbb{R}^{k_l \times n_l}$, COCA layer completes its functionality by using $\mathbf{\Pi}$ to pool and aggregate all output cluster features and attributes to the next layer. To enforce unsupervised feature learning throughout the COCA-Net hierarchy, we make use of skip connections \textit{across} layers, starting from layer $l=2$ up to the final layer $l=L$. In each layer $l \geq 2$, the inter-layer skip connection procedure involves merging the output cluster masks generated in the last two layers, specifically $\mathbf{\Pi}^{(l-1)}$ and $\mathbf{\Pi}^{l}$. This merging operation allows COCA-Net to skip-connect the feature tensors between every second layer, $\mathbf{X}^{(l-2)}$ and $\mathbf{X}^{(l)}$, facilitating unsupervised learning with a hierarchical deep learning model. Details on the aggregation and skip connection procedures are provided in Supplementary Material.

\subsection{COCA-Net}
\label{sec:cocanet}
Following a procedure similar to most sequential, agglomerative, hierarchical, and non-overlapping clustering methods \cite{ComprehensiveSurveyHAC}, each COCA layer partitions its inputs into non-overlapping windows, then performs clustering and pooling within each window in parallel to produce a reduced number of output clusters. This strategy is repeated at each hierarchical level, progressively reducing the number of input nodes to ultimately obtain object clusters at the last layer. The clustering assignments generated at each layer of the COCA-Net hierarchy are combined into a tree-structure called dendrogram, similar to most HAC algorithms. This dendrogram is used to evaluate the segmentation performance of COCA-Net's encoder sub-network.

\begin{figure*}[t!]
  \centering
  \fbox{\includegraphics[width=.8\textwidth,height=.5\textwidth]{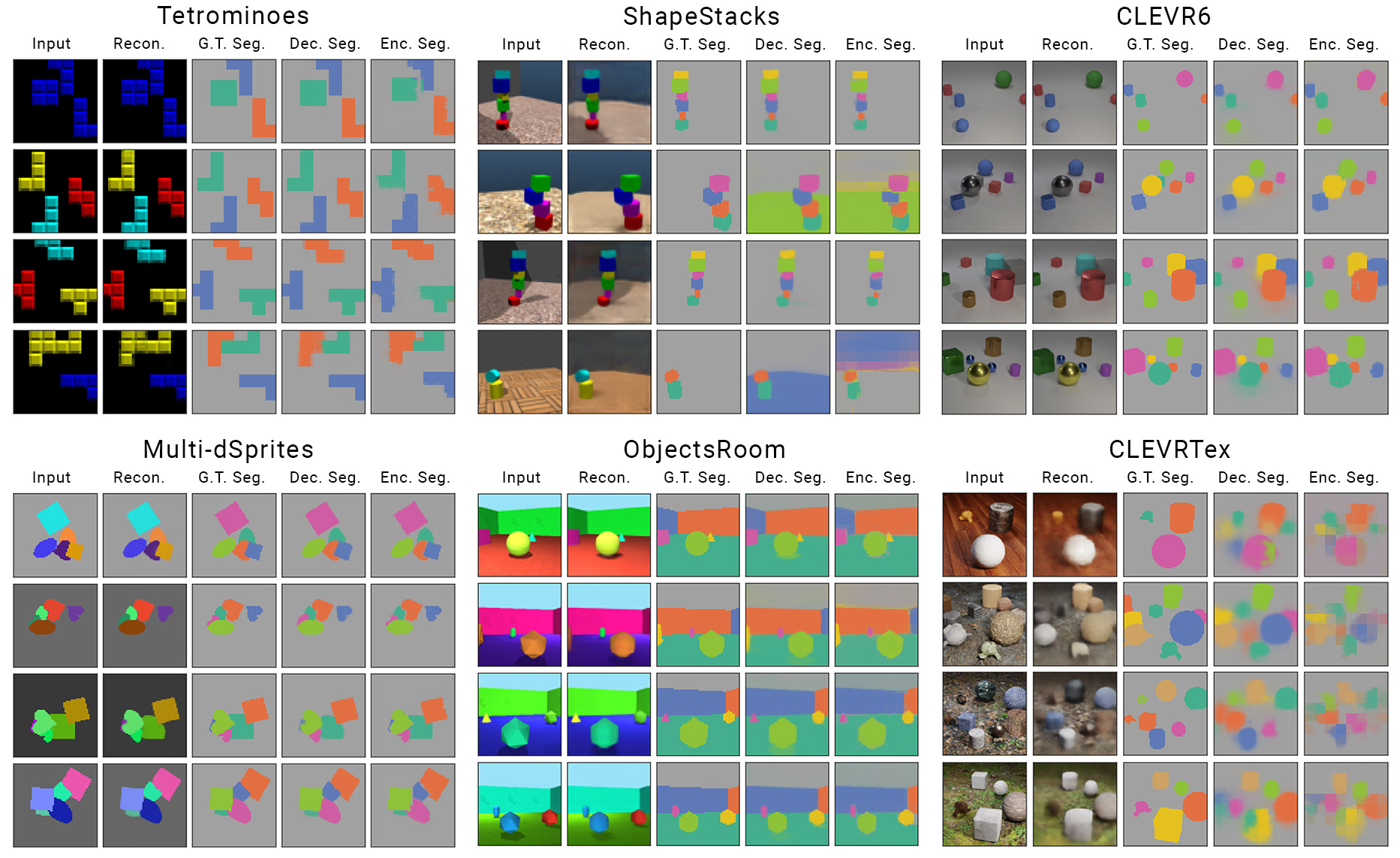}}
    \caption{Qualitative results of the COCA-Net architecture across six datasets examined in this work. For each dataset, we present four challenging samples along with COCA-Net’s reconstructions. Ground truth segmentation masks, as well as segmentation masks from the decoder and encoder sub-networks, are also included. The mask with the highest intersection with the background segment is shown in gray.}
    \label{fig:qualitative}
\end{figure*}
\section{Experiments and Results} 
\subsection{Baselines and Datasets} With the aim of establishing a comprehensive benchmark for our experiments, we utilize the OCL library provided by \cite{Dittadietal22}. We compare our architecture against three of the state-of-the-art object-centric models that generate scene segmentation masks in an unsupervised way: GEN-v2 \cite{GENESISV2_Engelcke2021}, INVSA \cite{invsa} and BOQSA \cite{boqsa}. The evaluation of these baselines and the proposed architecture is carried out across six widely adopted synthetic object discovery datasets, namely, Tetrominoes, Multi-dSprites, ObjectsRoom \cite{multiobjectdatasets19}, Shapestacks \cite{shapestacks}, CLEVR6 and CLEVRTex \cite{clevr}. 

To establish a fair comparison between different methods, we utilize the SBD decoder version of each method, as it can be considered as the standard practice for OCL methods on these datasets. Moreover, we optimize solely the pixel-reconstruction loss for all methods and train all architectures from scratch. In addition, we set the number of output slots equal to the maximum number of objects in a dataset for all methods. For the baseline methods, we implement the default hyper-parameters shared in their original work. Remaining training details for COCA-Net and the baselines can be found in Supplementary Material.

\subsection{Metrics and Evaluation Configurations} In line with previous work \cite{GENESISV2_Engelcke2021, boqsa}, we adopt two performance metrics namely Adjusted Rand Index (ARI) \cite{ARI_Hubert} and mean Segmentation Covering (mSC) \cite{GENESIS_Engelcke2019} for segmentation performance and Mean Squared Error (MSE) for reconstruction success. In accordance with the customs of the literature, we provide the segmentation performance of the object slot masks that are generated at the end of the decoder sub-networks. To assess COCA-Net's potential to be leveraged as a stand-alone unsupervised scene segmentation encoder, we also compare object slot masks that are obtained at the encoder sub-networks of the overall pipeline. Finally, to analyze COCA-Net's overall scene segmentation performance, we include background segments in our evaluation criteria. We trained all models using three random seeds and report results as mean $\pm$ standard deviation.

\begin{table}[ht]
  \centering
  \hspace*{-2.em}
  \caption{\footnotesize{Quantitative Results of proposed COCA-Net and baseline BOQSA on Tetrominoes and Multi-dSprites datasets. Evaluation configurations are: masks generated by encoder, (1) only foreground objects, (2) including background segments. Scores are reported as mean ± standard deviation for 3 seeds.}}
  \label{tbl:tetro}
 \hspace*{-1.5em}
  \footnotesize{
  \begin{tabular}{lllll}
    \toprule
    \multicolumn{1}{c}{} & \multicolumn{2}{c}{ENC-FG Only} & \multicolumn{2}{c}{ENC-BG Included}\\ 
    \midrule
    \multicolumn{1}{c}{Name} & \multicolumn{1}{c}{ARI$\uparrow$} & \multicolumn{1}{c}{mSC$\uparrow$} & \multicolumn{1}{c}{ARI$\uparrow$} & \multicolumn{1}{c}{mSC$\uparrow$} \\
    \midrule
    \multicolumn{5}{c}{Tetrominoes} \\
    \midrule 
    BOQSA \cite{boqsa} &  \multicolumn{1}{c}{0.60$\pm$0.03} & \multicolumn{1}{c}{0.42$\pm$0.01} & \multicolumn{1}{c}{0.29$\pm$0.02} & \multicolumn{1}{c}{0.47$\pm$0.00} \\    
    COCA-Net (ours) &  \multicolumn{1}{c}{\textbf{0.74$\pm$0.04}} & \multicolumn{1}{c}{\textbf{0.70$\pm$0.03}} & \multicolumn{1}{c}{\textbf{0.71$\pm$0.09}} & \multicolumn{1}{c}{\textbf{0.75$\pm$0.06}} \\
    \midrule
    \multicolumn{5}{c}{Multi-dSprites}\\
    \midrule
    BOQSA \cite{boqsa} &  \multicolumn{1}{c}{0.75$\pm$0.01} & \multicolumn{1}{c}{0.55$\pm$0.01} & \multicolumn{1}{c}{0.34$\pm$0.06} & \multicolumn{1}{c}{0.56$\pm$0.02} \\
    COCA-Net (ours) & \multicolumn{1}{c}{\textbf{0.96$\pm$0.01}} & \multicolumn{1}{c}{\textbf{0.95$\pm$0.01}} & \multicolumn{1}{c}{\textbf{0.98$\pm$0.00}} & \multicolumn{1}{c}{\textbf{0.96$\pm$0.00}} \\
    \bottomrule
  \end{tabular}}
  \vspace{-10pt}
\end{table}

\subsection{Results} Quantitative results are summarized in Tables \ref{tbl:tetro}, \ref{quantitative}, whereas qualitative results are displayed across Figure \ref{fig:qualitative}. Note that an extension of Table \ref{tbl:tetro} can be found in Supplementary Material. As it can be observed from the overall results, COCA-Net outperforms the state-of-the-art on all datasets across most metrics and displays highly competitive performance in others. Regarding the evaluation of foreground object masks produced by the decoder sub-network, COCA-Net achieves the highest ARI and mSC performance in almost all datasets, sharing the lead with \cite{invsa} in ObjectsRoom. Inspecting the foreground object masks generated by the encoder sub-networks, a dendrogram in our case, COCA-Net significantly surpasses its competitors, achieving an improvement of nearly thirty percent in both metrics on the ShapeStacks dataset. Notably, COCA-Net is more robust compared to its baselines in almost every dataset, exhibiting less variance across different training runs. 

Examining the segmentation results including the background segments, one can observe that COCA-Net is a better option for holistic scene segmentation, providing approximately thirty percent performance increase over \cite{boqsa} and \cite{invsa} on ObjectsRoom. The only exception is COCA-Net's ARI results on background segments of the ShapeStacks dataset. Both ShapeStacks and ObjectsRoom datasets contain multiple background segments but ShapeStacks dataset provides a single ground truth mask for all background segments. Therefore, COCA-Net's poor ARI performance on background segments of ShapeStacks is expected. Analyzing the qualitative results depicted in Figure \ref{fig:qualitative}, one can conclude that COCA-Net produces coherent segments that match with human intuition across all datasets, including ShapeStacks, where this intuitive behavior of COCA-Net is also supported by the quantitative mSC results. 

\begin{table*}[t]
  \footnotesize{
  \caption{\footnotesize{Unsupervised scene segmentation results of three baseline models and proposed COCA-Net on four multi-object datasets based on a total of nine performance evaluation metrics assessed across four different configurations. Scores are reported as mean $\pm$ standard deviation for 3 seeds. The four different evaluation configurations are: (1) masks generated by decoder, only foreground objects, (2) masks generated by decoder, with background segments, (3) masks generated by encoder only foreground objects, (4) masks generated by encoder with background segments.}}
  \label{quantitative}
  \centering
  \hspace*{1.em}
  \begin{tabular}{l l l l l l l l l l }
    \toprule
    & \multicolumn{2}{c}{DEC-FG Only} & \multicolumn{2}{c}{DEC-BG Included} & \multicolumn{2}{c}{ENC-FG Only} & \multicolumn{2}{c}{ENC-BG Included} \\
    \cmidrule(r){2-3}  \cmidrule(r){4-5}  \cmidrule(r){6-7}  \cmidrule(r){8-9}
    \multicolumn{1}{c}{Name} & \multicolumn{1}{c}{ARI$\uparrow$} & \multicolumn{1}{c}{mSC$\uparrow$}   &  \multicolumn{1}{c}{ARI$\uparrow$} &  \multicolumn{1}{c}{mSC$\uparrow$} &  \multicolumn{1}{c}{ARI$\uparrow$}   & \multicolumn{1}{c}{mSC$\uparrow$}  & \multicolumn{1}{c}{ARI$\uparrow$}   & \multicolumn{1}{c}{mSC$\uparrow$}      &  \multicolumn{1}{c}{MSE$\downarrow$} \\
    \midrule
    \multicolumn{10}{c}{ObjectsRoom} \\
    \midrule
    GEN-v2 \cite{GENESISV2_Engelcke2021}  & 0.86$\pm$0.01   &		0.57$\pm$0.03   &	0.11$\pm$0.00   &		0.38$\pm$0.01   &	0.67$\pm$0.07   &		0.18$\pm$0.02   &	0.13$\pm$0.00   &	 0.23$\pm$0.00   &	0.003$\pm$0.000 \\
    INV-SA \cite{invsa} &  \textbf{0.88$\pm$0.00}	&  0.80$\pm$0.01	&  0.66$\pm$0.12	&  0.68$\pm$0.07 &  0.80$\pm$0.02	&  0.44$\pm$0.02	&  0.36$\pm$0.01	&  0.39$\pm$0.01	&  0.001$\pm$0.001 \\  
    BOQ-SA \cite{boqsa} &  0.87$\pm$0.01	&  \textbf{0.83$\pm$0.00}	&  0.57$\pm$0.00	&  0.63$\pm$0.00	&  0.59$\pm$0.02	&  0.69$\pm$0.03	&  0.52$\pm$0.01	&  0.56$\pm$0.01	&  \textbf{0.001$\pm$0.000} \\
    COCA-Net (ours)   &  \textbf{0.88$\pm$0.00}    &  0.82$\pm$0.01	&  \textbf{0.95$\pm$0.01}	&  \textbf{0.87$\pm$0.02}	&  \textbf{0.87$\pm$0.01}	&  \textbf{0.82$\pm$0.01}	&  \textbf{0.94$\pm$0.02}	&  \textbf{0.86$\pm$0.02}	&  \textbf{0.001$\pm$0.000} \\
    \midrule
    \multicolumn{10}{c}{ShapeStacks} \\
    \midrule
    GEN-v2 \cite{GENESISV2_Engelcke2021} & 0.83$\pm$0.01  &	0.70$\pm$0.01  &	\textbf{0.87$\pm$0.00}  &	0.77$\pm$0.01  &	0.54$\pm$0.04  &	0.47$\pm$0.03  &	\textbf{0.60$\pm$0.04}  &	0.58$\pm$0.03  &	0.003$\pm$0.000  \\
    INV-SA\cite{invsa} & 0.65$\pm$0.13 & 0.64$\pm$0.07 & 0.12$\pm$0.02 & 0.55$\pm$0.05 & 0.55$\pm$0.11 & 0.37$\pm$0.04 & 0.09$\pm$0.01 & 0.36$\pm$0.04 & 0.004$\pm$0.002 \\    
    BOQ-SA \cite{boqsa} & 0.83$\pm$0.09	& 0.80$\pm$0.09	& 0.21$\pm$0.07	& 0.73$\pm$0.10	& 0.49$\pm$0.16	& 0.59$\pm$0.13	& 0.18$\pm$0.09	& 0.58$\pm$0.13	& \textbf{0.001$\pm$0.000} \\
    COCA-Net (ours) & \textbf{0.91$\pm$0.01} &	\textbf{0.85$\pm$0.01} &	0.31$\pm$0.09 &	\textbf{0.79$\pm$0.01} &	\textbf{0.82$\pm$0.02} &	\textbf{0.85$\pm$0.01} &	0.25$\pm$0.04 &	\textbf{0.78$\pm$0.01}  &	0.006$\pm$0.003 \\
    \midrule
    \multicolumn{10}{c}{CLEVR6} \\
    \midrule
    GEN-v2 \cite{GENESISV2_Engelcke2021} & 0.43$\pm$0.10	& 0.18$\pm$0.00	& 0.08$\pm$0.01	& 0.21$\pm$0.00	& 0.17$\pm$0.24 & 0.07$\pm$0.05	& 0.05$\pm$0.07	& 0.16$\pm$0.01	& 0.004$\pm$0.000 \\
    INV-SA \cite{invsa}   &  0.96$\pm$0.01	& \textbf{0.87$\pm$0.03} &0.74$\pm$0.35 &\textbf{0.87$\pm$0.07}  &0.64$\pm$0.04 &0.54$\pm$0.02 &0.30$\pm$0.07 &0.55$\pm$0.04 &\textbf{0.000$\pm$0.000} \\
    BOQ-SA \cite{boqsa}   & 0.86$\pm$0.21   &	0.34$\pm$0.12 & 0.08$\pm$0.07    &  	0.33$\pm$0.11  & 0.78$\pm$0.26   	&0.30$\pm$0.12 &   0.08$\pm$0.07  &  0.31$\pm$0.13 & \textbf{0.000$\pm$0.000} \\   			   			   		   	   
    COCA-Net (ours) & \textbf{0.97$\pm$0.02}	&0.82$\pm$0.06	& \textbf{0.88$\pm$0.04}	&0.85$\pm$0.05& \textbf{0.96$\pm$0.03}	& \textbf{0.76$\pm$0.02} & \textbf{0.82$\pm$0.08}	& \textbf{0.79$\pm$0.02}	& \textbf{0.000$\pm$0.000} \\
    \midrule
    \multicolumn{10}{c}{CLEVRTex} \\
    \midrule
    GEN-v2 \cite{GENESISV2_Engelcke2021} & 0.22$\pm$0.08 &	0.10$\pm$0.00 &	0.02$\pm$0.04 &	0.14$\pm$0.00 &	0.00$\pm$0.00 &	0.04$\pm$0.00 &	0.00$\pm$0.00 &	0.15$\pm$0.00 &	0.009$\pm$0.000 \\    
    INV-SA \cite{invsa}   & 0.60$\pm$0.12 &	0.35$\pm$0.13 &	0.21$\pm$0.11 &	0.38$\pm$0.12 &	0.53$\pm$0.11 &	0.28$\pm$0.08 &	0.14$\pm$0.05 &	0.31$\pm$0.06 &	0.009$\pm$0.003 \\
    BOQ-SA \cite{boqsa}  &  0.69$\pm$0.06	&  0.44$\pm$0.07	&  0.18$\pm$0.05	&  0.43$\pm$0.07	&  \textbf{0.53$\pm$0.03}	&  0.34$\pm$0.07	&  0.14$\pm$0.06	&  0.34$\pm$0.08	&  \textbf{0.005$\pm$0.002} \\
    COCA-Net (ours) & \textbf{0.76$\pm$0.02}  &	\textbf{0.54$\pm$0.03}  &	\textbf{0.50$\pm$0.06}  &	\textbf{0.57$\pm$0.03}  &	0.44$\pm$0.03  &	\textbf{0.39$\pm$0.03}  &	\textbf{0.39$\pm$0.03}  &	\textbf{0.45$\pm$0.03}  &	0.008$\pm$0.001 \\
    \bottomrule
  \end{tabular}}
\end{table*}

\subsection{Ablation Studies} 
\label{sec:ablation}
We conduct two ablation studies for our design choices across the COCA-Net architecture. These ablation studies are designed as follows; (1) to verify the effectiveness of compactness based mask selection and SBC based clustering, we consider the original recipe laid out in GEN-v2 \cite{GENESISV2_Engelcke2021}, where anchor node selection is carried out by sampling a random pixel from a uniform distribution that is concealed by the current scope. We name this version as COCA-Net-RAS to indicate ``random anchor node selection". (2) To assess COCA-Net's flexibility on generating a varying number of output slots, we develop a version of COCA-Net that keeps on generating object masks until a stopping condition is met, e.g., number of remaining elements in the scope $\mathbf{Z}$ drops below a threshold, similar to one described in \cite{GENESISV2_Engelcke2021}. This version, that we refer to as COCA-Net-Dyna, is trained on CLEVR6 with a fixed number of output slots and evaluated on CLEVR10 with dynamic number of output slots.  Results for these ablation studies are displayed in Table \ref{tbl:ablation}. The integration of compactness scoring significantly enhances the performance of the COCA layer in multi-object segmentation. Furthermore, COCA-Net demonstrates the ability to dynamically adjust its output representation capacity without compromising performance.

\vspace{-15pt}
\begin{table}[ht]
  \centering
  \hspace*{-2.em}
\caption{\footnotesize{Results of the Ablation Studies (ASs) conducted for the proposed COCA-Net architecture. For both studies, we use decoder-generated foreground segmentation accuracy as the comparison criterion. }}
  \label{tbl:ablation}
  \vspace{5pt}
  \footnotesize{
  \begin{tabular}{lllll}
    \toprule
    \multicolumn{5}{l}{\textbf{AS 1: Effectiveness of Compactness Measurement}} \\
    \midrule
    \multicolumn{1}{c}{Model} & \multicolumn{2}{c}{ARI} & \multicolumn{2}{c}{mSC}\\ 
    \midrule
    COCA-Net on ObjectsRoom &  \multicolumn{2}{c}{\textbf{0.894}} & \multicolumn{2}{c}{\textbf{0.832}} \\
    COCA-Net-RAS on ObjectsRoom &  \multicolumn{2}{c}{0.832} & \multicolumn{2}{c}{0.739} \\
    \midrule
    \multicolumn{5}{l}{\textbf{AS 2: Variable Number of Output Slots}}\\
    \midrule
    \multicolumn{1}{c}{Model} & \multicolumn{2}{c}{ARI} & \multicolumn{2}{c}{mSC}\\ 
    \midrule
    COCA-Net on CLEVR6 & \multicolumn{2}{c}{0.985} & \multicolumn{2}{c}{0.881} \\
    COCA-Net-Dyna on CLEVR10 & \multicolumn{2}{c}{0.978} & \multicolumn{2}{c}{0.840}\\
    \bottomrule
  \end{tabular}}
\end{table}

\vspace{-10pt}
\section{Conclusion}
\vspace{-4pt}

 COCA-Net provides state-of-the-art segmentation results across various configurations and datasets. Notably, it is more robust compared to its competitors, yielding significantly lower variance across training runs. Segmentation masks generated by COCA-Net, both in the encoder and decoder sub-networks of its pipeline, exhibit accurate and highly competitive performance, highlighting its potential as a stand-alone unsupervised image segmentation encoder. In addition, COCA-Net can produce more coherent background segments compared to its baselines and can dynamically adjust the number of output slots. 
 
 A current shortcoming of COCA-Net is its bottom-to-top clustering approach, which lacks top-down feedback. This prevents the model from correcting errors made in earlier clustering layers. Future work might consider implementing a top-down feedback pathway, possibly replacing the standard SBD decoder architecture. Another drawback of the COCA layer, as currently framed, is the calculation of moment of inertia of the reference circle. This mass-normalized compactness measurement involves pairwise comparison of densities within each affinity mask, creating a heavy memory requirement. Future work can address this shortcoming by implementing an efficient approximation for the MI calculation of the reference shape.
{
    \small
    \bibliographystyle{ieeenat_fullname}
    \bibliography{main}
}

\clearpage
\maketitlesupplementary
We provide a complexity analysis of our method in Section \ref{sec:complex}. Subsequently, we present architectural details in Section \ref{sec:suppl_imp}, organized similarly to Section \ref{sec:method_3_0} of the main paper. Implementation details for COCA-Net and the baselines are provided in Section \ref{sec:suppl_training}. Additional results, including initial experiments on real-world datasets \ref{sec:suppl_init_real}, additional quantitative \ref{sec:suppl_quant} and qualitative \ref{sec:suppl_qual} results are presented throughout Section \ref{sec:suppl_add_results}. Extensions of the results from our ablation studies are also included in Section \ref{sec:suppl_abl}.

\section{Complexity Analysis}
\label{sec:complex}
Suppose that COCA-Net is to process a feature image consisting of $N\times N$ pixels. At each layer, this feature image is partitioned into $U \times U$ non-overlapping windows. The complexity burden of a COCA layer resides on the generation of affinity masks and their compactness measurement. In the worst case, COCA generates an affinity mask per node,  which introduces $O(U^2\cdot U^2)$ operations for the overall affinity masks generation process. In the first layer, we have a total of $\frac{N}{U}\cdot\frac{N}{U}=\frac{N^2}{U^2}$ windows, hence the complexity is $O(U^2\cdot N^2)$ for affinity masks generation. Our compactness measurement involves finding the node pair with minimum density in each affinity mask (see Eq.~\ref{eq:compactness_of_affinities}). The complexity associated with this operation increases above affinity mask generation, resulting in a complexity of $O(U^2\cdot U^3)$. Therefore, the first layer of COCA has a complexity of $O(U^3\cdot N^2)$. At the second layer, we repeat the same partitioning strategy to generate affinity masks and measure their compactness. Hence, we again need $O(U^2\cdot U^3)$ operations within a window and we have $\frac{N}{U^2}\cdot\frac{N}{U^2}=\frac{N^2}{U^4}$ of these windows at the second layer. Therefore, the overall complexity for this layer is $O(U\cdot N^2)$. Similarly at the third layer, we have a complexity of $O(\frac{N^2}{U})$ and so on. 

In total, we have $\log_UN $ layers, where $N$ is a power of $U$. Assuming that $U$ is kept constant and small, e.g., $U=4$, throughout the hierarchy, we have $O(N^2)$ complexity for each layer. Altogether, these $\log_UN $ layers add up to $O(N^2\cdot \log N)$ complexity. Comparing to our baselines, the complexity of  SA (slot attention) model \cite{sa} can be considered as $O(N^2)$. SA conducts its query normalized cross attention operation on the full resolution feature image, which has a complexity of $O(N^2)$. On the contrary, instead of operating on this large $N\times N$ image directly, COCA-Net operates on $U\times U$ windows and handles $\frac{N^2}{U^2}$ of windows in parallel. This structure makes COCA-Net inherently suitable for parallel implementation, as the operations within each window are independent. It should be noted that the number of elements that COCA-Net processes decreases by a factor of $U^2$ at each layer, hence it has the potential to be scaled up to deeper hierarchies and address object discovery in higher resolution images. 

\section{Architectural Details}
\label{sec:suppl_imp}
\subsection{Pixel Feature Encoder}
We use a simple backbone to initialize pixel features that encode both appearance and positional information. Specifically, the input image is first processed through a single convolutional layer that preserves resolution to generate appearance-based features. Positional information for each pixel is then incorporated through a positional encoding scheme borrowed from \cite{sa}, which normalizes the initial pixel coordinates in four cardinal directions and learns a projection that maps these 4-dimensional position vectors to the output feature dimensions of the convolutional layer. After summing the positional embedding with the convolutional features of the pixels, we apply a GroupNorm normalization across features and finally use a Multi-Layer Perceptron (MLP) with a single hidden layer to fuse the position and appearance information. The resulting tensor forms the initial image features $\mathbf{X}^{0} \in \mathbb{R}^{h_0\times w_0\times d_0}$, that we feed into our COCA-Net hierarchy at layer $l=1$.

\subsection{COCA Layer (Section \ref{subsec:coca})}
COCA layer can be viewed as a clustering variant of ViT-22B. It begins with a stack of original ViT-22B layers to refine the features of input nodes and then continues as a clustering variant of ViT-22B. While the overall structure of COCA resembles a ViT-22B, it differs significantly. 

Rather than relating two distinct projections—query and key projections to compute attention weights as in the original self-attention formulation—we project the representations of input nodes onto the same high-dimensional space to compute affinities, e.g, using only query projection. In addition, we use these affinity masks to identify clusters among input nodes and apply feature pooling based on the masks of these clusters, unlike ViT-22B, which maintains the same number of nodes from input to output across its layers.

Apart from these fundamental differences, most of the remaining architecture resembles a ViT-22B, as also illustrated in Figure \ref{fig:short-b}, we also learn a parallel FFN, a values projection, an output projection and employ pre-norm skip connection from the features that are just partitioned into non-overlapping windows.  

\begin{figure*}[t!]
  \centering
   \fbox{\includegraphics[width=.8\textwidth,height=.35\textwidth]{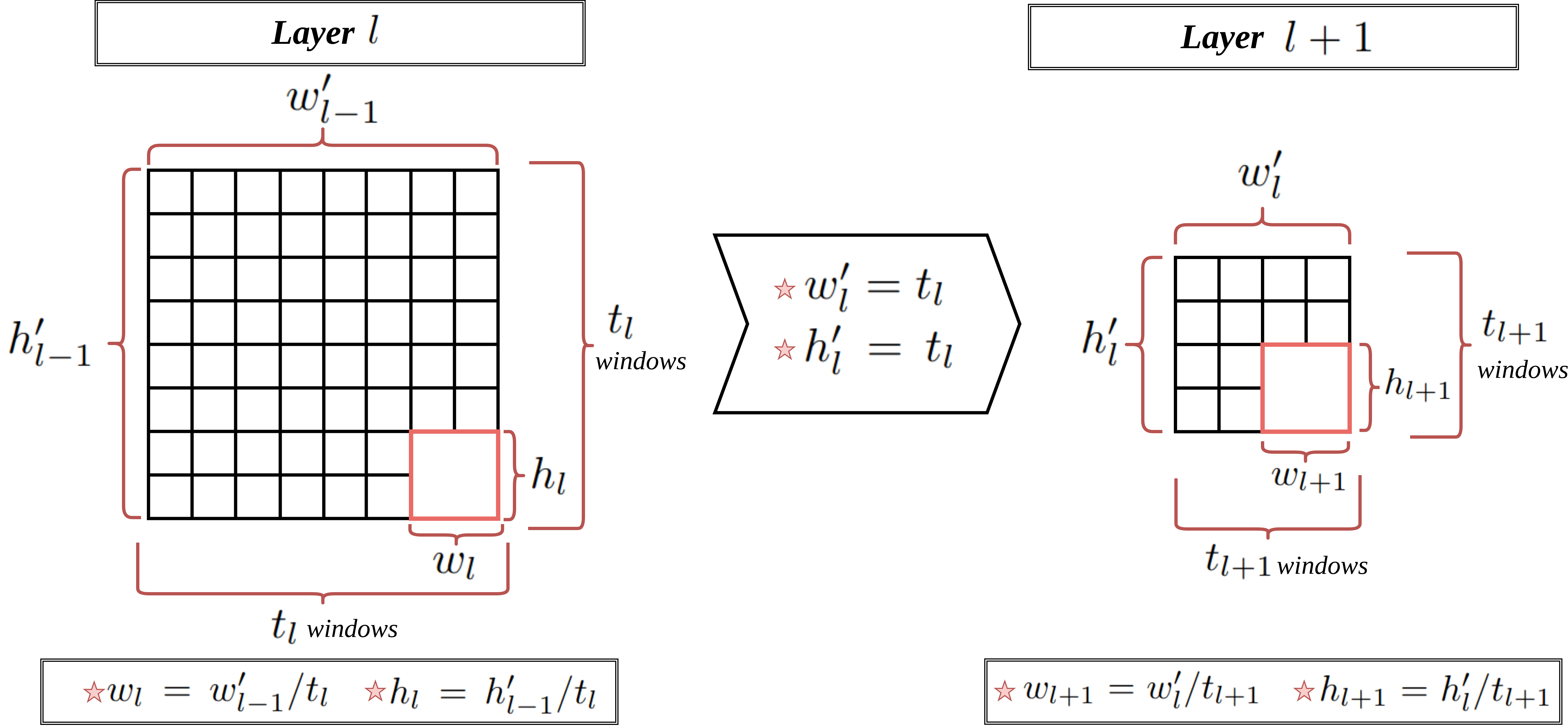}}
    \caption{Visualization of the Non-Overlapping partitioning strategy. }
    \label{fig:non_overlapping}
\end{figure*}

\subsubsection{Feature Refiner ViT-22B (Section \ref{sec:feature_refinement})}
A single ViT-22B layer \cite{vit22b_dehghani2023} in COCA adopts Multi-Head Self-Attention (MHSA) \cite{Attention_Vaswani2017} in parallel with the Feed Forward Network (FFN) in a Pre-Norm skip connection configuration. For the details, reader can refer to the original paper \cite{vit22b_dehghani2023}. In COCA, the only modification to the original recipe is that we leverage GroupNorm operations over LayerNorm, where the former is often preferred over the latter in computer vision architectures.

The number of ViT-22B layers in the feature refinement stack of a single COCA layer may vary depending on the COCA layer’s position in the hierarchy. Specifically, we adopt an increasing number of ViT-22B layers as we move up through the hierarchy. For instance, the first COCA layer contains a single ViT-22B layer in its feature refinement stack, whereas the second COCA layer includes two ViT-22B layers, and so on.

The spatial extent of the ViT-22B layers’ operations also evolves throughout the hierarchy. In the earlier layers, the attention context is constrained to smaller windows (e.g., a kernel size of $3 \times 3$), enabling feature refinement to focus on localized neighborhoods. Conversely, in the final layers, the attention context expands to a global scale, facilitating the integration of broader contextual information. Throughout this process, the stack of ViT-22B layers preserves the cardinality of the input node set; that is, a query is generated for every node, and an appropriate attention context is constructed within the spatial neighborhood for each query.

\subsubsection{Partitioning Hierarchy (Section \ref{sec:non_overlapping_partitioning})}
\label{sec:suppl_partition}

A visual illustration of the non-overlapping windows partitioning strategy is shared in Figure \ref{fig:non_overlapping}. After feature refinement, at each layer $l$, COCA contains an input node set with $h_{l-1}^\prime\times w_{l-1}^\prime \times k_{l-1}$ elements. COCA partitions this input node set into non-overlapping windows, where each window has $h_{l}\times w_{l}\times k_{l-1}$
nodes and there are a total of $t^2_l$ of such windows. In other words, input nodes are divided into $t_l$ groups along both spatial axes, i.e., $h_l=h_{l-1}^\prime/t_l$ and $w_l=w_{l-1}^\prime/t_l$ \footnote{For notational simplicity, we assume that the input image is square rather than rectangle, thus we do not need $t^h_l$ and $t^w_l$}. Note that $k_0=1$ and $h^\prime_0=w^\prime_0$ are equal to input image resolution.

With this partitioning strategy, COCA operates on a total of $n_l=h_l \times w_l \times k_{l-1}$ input nodes in each window, then proceeds with affinity masks generation, compactness scoring and sequential clustering. After clustering is completed, COCA has produced an output clusters set comprising $1 \times 1 \times k_l$ elements. Thus, effectively collapsing spatial dimensions of the each window to obtain $k_l$ output clusters per window. This means that, for the next COCA layer at layer $l+1$, the input node set will contain $t_l \times t_l \times k_l$ elements, i.e., $h_l^\prime=t_l$ and $w_l^\prime=t_l$. This COCA layer behaves in the exact same manner, first partitions this input set of $h_{l}^\prime\times w_{l}^\prime \times k_{l}$ elements into $t_{l+1}^2$ windows, where each window contains $h_{l+1}\times w_{l+1} \times k_l$ nodes. 

\subsubsection{Affinity Masks Generation (Section \ref{sec:affinity_masks})}
\label{sec:supp_affinity}
Inspired by ViT-22B, we employ pre-norm skip connections within a COCA layer, hence, the input refined and partitioned features tensor $\mathbf{X}^{l} \in \mathbb{R}^{n_l \times d_l}$ is used for skip connection with pooling after clustering is completed. Next, we apply the GroupNorm normalization to this features tensor and then use it  as input to three learnable projections, two linear and one non-linear; queries projection $\mathbf{Q}^l \in \mathbb{R}^{d_l \times d_l}$, values projection $\mathbf{V}^l \in \mathbb{R}^{d_l \times d_l}$ and the parallel multi-layer perceptron $\mathbf{MLP}^l_1 \in \mathbb{R}^{d_l \times d_l}$, $\mathbf{MLP}^l_2 \in \mathbb{R}^{d_l \times d_l}$:
\begin{subequations}
\begin{align}
 \hat{\mathrm{X}}^{l}_{ij} &= \mathrm{GroupNorm}_{j} \left( \mathrm{X}^{l}_{ij} \right) \label{eqn:supp_affn_line-1}\\
    \hat{\mathbf{Q}}^{l}_i &= \hat{\mathbf{X}}^{l}_i \mathbf{Q}^{l}\label{eqn:supp_affn_line-2}\\
     \hat{\mathbf{V}}^{l}_i &= \hat{\mathbf{X}}^{l}_i \mathbf{V}^{l}\label{eqn:supp_affn_line-3}\\
      \hat{\mathbf{\Psi}}^{l}_i &= (\mathrm{GELU}(\hat{\mathbf{X}}^{l}_i \mathbf{MLP}_1^{l})) \mathbf{MLP}_2^{l}\label{eqn:supp_affn_line-4}
\end{align}
\end{subequations}
where $\mathrm{GroupNorm}_{j}$ indicates that normalization is performed across the last axis of the features tensor. We use the query projected features as our common high-dimensional space to compute affinities on, whereas the value projected features and the features embedded by the parallel MLP will be used after generating cluster assignments (See Fig. \ref{fig:short-b}). 

In order to compute affinities between each pair of nodes within a window, we apply GroupNorm normalization to query projected features:
\begin{equation}
    \mathrm{Y}^{l}_{ij} = \mathrm{GroupNorm}_{j}( \hat{\mathrm{Q}}^{l}_{ij})
\end{equation}
where $\mathbf{Y}^{l} \in \mathbb{R}^{n_l \times d_l}$ denotes the projected and normalized feature vectors of all nodes. Following ViT-22B, we do not learn any parameters for the normalization operations involved throughout COCA layer. After all node features are projected to the same high-dimensional space and normalized, to produce the affinity masks, we measure Euclidean distances between features of each node pair, as given in Eq. \ref{eq:euc_distances} of the main paper. 

This naive computation of $\mathbf{E}^{l} \in \mathbb{R}^{n_l\times n_l}$ uses ``all-to-all" affinities to generate candidate masks, however, to improve efficiency, we can optionally reduce the density of the affinity masks using a spatially dilated sampling strategy, producing sparser masks with size $u_l \times n_l$ where $u_l << n_l$. As also mentioned in the main paper, we use soft-argmin to convert these distances to dynamic and adaptive affinities and employ min-max scaling to map these affinities to a proper range, i.e. in $[0,1]$ for compactness computation and SBC clustering:
\begin{subequations}
\begin{align}
    \tilde{\mathrm{\Lambda}}^{l}_{ij} &= \mathrm{soft{\text-}argmin}_{j} \left( \mathrm{E}^{l}_{ij} \right) \label{eqn:supp_affn_line-5}\\
    \mathrm{\Lambda}^{l}_{ij} &= \frac{\tilde{\mathrm{\Lambda}}^{l}_{ij}-\min_j(\tilde{\mathrm{\Lambda}}^{l}_{ij})}{\max_j(\tilde{\mathrm{\Lambda}}^{l}_{ij})-\min_j(\tilde{\mathrm{\Lambda}}^{l}_{ij})}\label{eqn:supp_affn_line-6}
\end{align}
\end{subequations}

\subsubsection{Compactness Scoring (Section \ref{sec:compactness})}
Recall that a single affinity mask $\boldsymbol{\Lambda}^l_i$, can be considered as a flattened two dimensional grid of affinities that constitute a non-uniform density shape and that we utilize five physical attributes for each node throughout the hierarchy, i.e., area $\mathbf{A}^{l} \in \mathbb{R}^{n_l\times1}$, mass $\mathbf{M}^{l} \in \mathbb{R}^{n_l\times1}$, density $\mathbf{D}^{l} \in \mathbb{R}^{n_l\times1}$, moment of inertia $\mathbf{I}^{l} \in \mathbb{R}^{n_l\times1}$ and mean position in the original image resolution $\mathbf{P}^{l} \in \mathbb{R}^{n_l \times 2}$. The positions of each input node to the first layer $\mathbf{P}^{1}_{i}$ are initialized with the corresponding pixel coordinates, while the area $\mathrm{A}^{1}_{i}$, mass $\mathrm{M}^{1}_{i}$, and density $\mathrm{D}^{1}_{i}$ of nodes are set to ones. In line with \cite{EfficientMesCompactness_Li2013}, the moment of inertia $\mathrm{I}^{1}_{i}$ for each pixel is set to $1/6$. 

As noted in Section \ref{sec:compactness}, we compute the intermediate attributes by first broadcasting then computing an element-wise product with the affinity masks $\mathbf{\Lambda}^{l} \in \mathbb{R}^{n_l \times n_l}$:
\begin{subequations}
\begin{align}
\tilde{\mathbf{A}}^{l} &= \mathbf{1}^{n_l}(\mathbf{A}^l)^\intercal \odot \mathbf{\Lambda}^{l}\label{eqn:supp_comp_line-1} \\
\tilde{\mathbf{D}}^{l} &= \mathbf{1}^{n_l}(\mathbf{D}^l)^\intercal \odot \mathbf{\Lambda}^{l} \label{eqn:supp_comp_line-2}\\
\tilde{\mathbf{I}}^{l} &= \mathbf{1}^{n_l}(\mathbf{I}^l)^\intercal \odot \mathbf{\Lambda}^{l} \label{eqn:supp_comp_line-3} \\
\tilde{\mathbf{M}}^{l} &= \tilde{\mathbf{A}}^{l} \odot \tilde{\mathbf{D}}^{l}\label{eqn:supp_comp_line-4}
\end{align}
\label{eqn:all-lines}
\end{subequations} 
where $\mathbf{1}^{n_l}$ is a column vector of ones of size $n_l$, $\odot$ denotes the Hadamard product and $\tilde{\mathbf{A}}^{l}, \tilde{\mathbf{D}}^{l}, \tilde{\mathbf{I}}^{l}, \tilde{\mathbf{M}}^{l} \in \mathbb{R}^{n_l \times n_l}$. Notably, the mass attribute is effectively scaled with the square of the affinities (See Eq. \ref{eqn:supp_comp_line-4}). Empirically, we found that this formulation discourages uncertain affinities in output cluster masks (e.g., an affinity of 0.4 is scaled to 0.16), while affinity values close to 1 remain only modestly affected. After intermediate attributes computation, we compute the compactness scores $\mathbf{C}^l \in \mathbb{R}^{n_l}$ of affinity masks using Eq. \ref{eq:compactness_of_affinities} in the main paper.

\begin{table*}[t]
  \centering
  \footnotesize{
  \caption{\footnotesize{Hyper-parameters for Pixel Feature Encoder Backbone across six datasets}}
  \label{tbl:hyperparameter_backbone}
  \begin{tabular}{l l l l l l l}
    \toprule
    \multicolumn{1}{c}{Dataset} & \multicolumn{1}{c}{Channels} &  \multicolumn{1}{c}{Pos. Emb. Channels} & 
    \multicolumn{1}{c}{MLP Out Channels} &
    \multicolumn{1}{c}{Kernel Size}   &  \multicolumn{1}{c}{Stride} &  \multicolumn{1}{c}{Padding} \\
    \midrule 
\multicolumn{1}{c}{Tetrominoes} & \multicolumn{1}{c}{32} & \multicolumn{1}{c}{32} & \multicolumn{1}{c}{64} & \multicolumn{1}{c}{$3\times3$}   &  \multicolumn{1}{c}{$1\times1$} &  \multicolumn{1}{c}{$1\times1$} \\

\multicolumn{1}{c}{Multi-dSprites} & \multicolumn{1}{c}{64} & \multicolumn{1}{c}{64} & \multicolumn{1}{c}{64} & \multicolumn{1}{c}{$3\times3$}   &  \multicolumn{1}{c}{$1\times1$} &  \multicolumn{1}{c}{$1\times1$} \\

\multicolumn{1}{c}{ShapeStacks} & \multicolumn{1}{c}{64} & \multicolumn{1}{c}{64} & \multicolumn{1}{c}{64} & \multicolumn{1}{c}{$3\times3$}   &  \multicolumn{1}{c}{$1\times1$} &  \multicolumn{1}{c}{$1\times1$} \\

\multicolumn{1}{c}{ObjectsRoom} & \multicolumn{1}{c}{64}& \multicolumn{1}{c}{64} & \multicolumn{1}{c}{64} & \multicolumn{1}{c}{$3\times3$}   &  \multicolumn{1}{c}{$1\times1$} &  \multicolumn{1}{c}{$1\times1$} \\

\multicolumn{1}{c}{CLEVR6} & \multicolumn{1}{c}{96} & \multicolumn{1}{c}{96}  & \multicolumn{1}{c}{96} & \multicolumn{1}{c}{$5\times5$}   &  \multicolumn{1}{c}{$1\times1$} &  \multicolumn{1}{c}{$2\times2$} \\

\multicolumn{1}{c}{CLEVRTex} & \multicolumn{1}{c}{128} & \multicolumn{1}{c}{128} & \multicolumn{1}{c}{96} &\multicolumn{1}{c}{$5\times5$}   &  \multicolumn{1}{c}{$1\times1$} &  \multicolumn{1}{c}{$2\times2$} \\
    \bottomrule
  \end{tabular}}
\end{table*}

\begin{table*}[t]
  \centering
  \footnotesize{  \caption{\footnotesize{Hyper-parameters for Spatial Broadcast Decoder employed in COCA-Net, across six datasets}}
\label{tbl:hyperparameter_sbd}
\hspace*{-3.5em}
  \begin{tabular}{l l l l l l l l}
    \toprule
    \multicolumn{1}{c}{Dataset} &  \multicolumn{1}{c}{Dataset Resolution} & \multicolumn{1}{c}{Broadcast Resolution} & \multicolumn{1}{c}{Channels}   &  \multicolumn{1}{c}{Kernel Sizes} &  \multicolumn{1}{c}{Strides} &  \multicolumn{1}{c}{Paddings}   & \multicolumn{1}{c}{Output Paddings} \\
    \midrule
    \multicolumn{1}{c}{Tetrominoes} &  \multicolumn{1}{c}{$32\times32$} & \multicolumn{1}{c}{$32\times32$} &  \multicolumn{1}{c}{$[32, 32, 32, 4]$}   &  \multicolumn{1}{c}{$[5, 5, 5, 3]$} &  \multicolumn{1}{c}{$[1, 1, 1, 1]$} &  \multicolumn{1}{c}{$[2, 2, 2, 1]$}   & \multicolumn{1}{c}{[0, 0, 0, 0]} \\   

     \multicolumn{1}{c}{Multi-dSprites} &  \multicolumn{1}{c}{$64\times64$} & \multicolumn{1}{c}{$64\times64$} & \multicolumn{1}{c}{$[32, 32, 32, 4]$}   &  \multicolumn{1}{c}{$[5, 5, 5, 3]$} &  \multicolumn{1}{c}{$[1, 1, 1, 1]$} &  \multicolumn{1}{c}{$[2, 2, 2, 1]$}   & \multicolumn{1}{c}{[0, 0, 0, 0]} \\    
    
    \multicolumn{1}{c}{ShapeStacks} &  \multicolumn{1}{c}{$64\times64$} & \multicolumn{1}{c}{$64\times64$} & \multicolumn{1}{c}{$[32, 32, 32, 4]$}   &  \multicolumn{1}{c}{$[5, 5, 5, 3]$} &  \multicolumn{1}{c}{$[1, 1, 1, 1]$} &  \multicolumn{1}{c}{$[2, 2, 2, 1]$}   & \multicolumn{1}{c}{[0, 0, 0, 0]} \\    
    
    \multicolumn{1}{c}{ObjectsRoom} &  \multicolumn{1}{c}{$64\times64$} & \multicolumn{1}{c}{$64\times64$} & \multicolumn{1}{c}{$[32, 32, 32, 4]$}   &  \multicolumn{1}{c}{$[5, 5, 5, 3]$} &  \multicolumn{1}{c}{$[1, 1, 1, 1]$} &  \multicolumn{1}{c}{$[2, 2, 2, 1]$}   & \multicolumn{1}{c}{[0, 0, 0, 0]} \\    
    
    \multicolumn{1}{c}{CLEVR6} &  \multicolumn{1}{c}{$128\times128$} & \multicolumn{1}{c}{$8\times8$} & \multicolumn{1}{c}{$[48, 48, 48, 48, 48, 4]$}   &  \multicolumn{1}{c}{$[5, 5, 5, 5, 5, 3]$} &  \multicolumn{1}{c}{$[2, 2, 2, 2, 1, 1]$} &  \multicolumn{1}{c}{$[2, 2, 2, 2, 2, 1]$}   & \multicolumn{1}{c}{$[1, 1, 1, 1, 0, 0]$} \\

     \multicolumn{1}{c}{CLEVRTex} &  \multicolumn{1}{c}{$128\times128$} & \multicolumn{1}{c}{$8\times8$} & \multicolumn{1}{c}{$[64, 64, 64, 64, 64, 4]$}   &  \multicolumn{1}{c}{$[5, 5, 5, 5, 5, 3]$} &  \multicolumn{1}{c}{$[2, 2, 2, 2, 1, 1]$} &  \multicolumn{1}{c}{$[2, 2, 2, 2, 2, 1]$}   & \multicolumn{1}{c}{$[1, 1, 1, 1, 0, 0]$} \\
    
\bottomrule
  \end{tabular}}
\end{table*}

\subsubsection{Sequential Clustering (Section \ref{sec:seq_clus})} 
A single cluster mask generation scheme in a COCA layer was discussed in detail in Section \ref{sec:seq_clus} of the main paper. Here, with Algorithm \ref{alg:1}, we provide the pseudo-code for the overall compactness-based sequential clustering algorithm.

\begin{algorithm}
   \SetKwInput{KwData}{Input}
   \SetKwInput{KwResult}{Output}
  \KwData{Compactness Scores $\mathbf{C}^l \in \mathbb{R}^{n_l}$, Affinity Masks $\mathbf{\Lambda}^{l} \in \mathbb{R}^{n_l\times n_l}$}
  \KwResult{Cluster Masks $\mathbf{\Pi}^{l} \in \mathbb{R}^{k_l\times n_l}$ }
  \textbf{Initialise:} Cluster Masks $\mathbf{\Pi}^{l} = \varnothing $, Scope $\mathbf{Z}^l = \mathbf{1}^{n_l}$\\
  \While{not stopping condition$(\mathbf{Z}^l)$}{
    $\mathbf{C}^{l} = \mathbf{C}^{l} \odot \mathbf{Z}^l$\\
    $\omega = \mathrm{argmax}_i(\mathrm{C}_i^{l}) $\\
    $\mathbf{\Pi}^{l}.\mathrm{append}(\mathbf{\Lambda}_\omega^{l} \odot \mathbf{Z}^l)$\\
    $\mathbf{Z}^l = \mathbf{Z}^l \odot (1-\mathbf{\Pi}_{-1}^{l})$
  }
  $\mathbf{\Pi}^{l}.\mathrm{append}(\mathbf{Z}^l)$
  \caption{Sequential Clustering}
  \label{alg:1}
\end{algorithm}
Note that in line 2 of Alg.\ref{alg:1}, the `stopping condition' for the original COCA-Net is met when the maximum number of objects in the dataset has been generated. For COCA-Net-Dyna, this condition is satisfied when a stopping condition is met, e.g., the sum of the scope variable drops below 2.5\% of its total size. In line 3, we first erode the compactness scores of nodes with an element-wise multiplication with the current scope. Next, in line 4, we find the most compact node among the remaining ones. With line 5, we first fetch the affinity mask that belongs to the most compact node, conceal it with the current scope and then add it to the list of generated cluster masks. We then update our scope, with line 6, to discard the nodes that are explained in this iteration. After the stopping condition is met, and the loop is finished, we finally add the remaining scope to our cluster masks list and complete the algorithm.

\begin{table*}[t]
  \centering
  \footnotesize{
  \caption{\footnotesize{Hyper-parameters for COCA-Net encoder across six datasets. Here, Q Kernel, Q Stride and Q Pad denote the parameters of unfolding operation applied on the queries of ViT-22B layer. Similarly, K Kernel, K Stride and K Pad stand for unfolding parameters for keys.}}
  \label{tbl:hyperparameter_cocanet}
  \hspace*{-5.em}
  \begin{tabular}{l l l l l l l l l l l l l}
    \toprule
    \multicolumn{1}{c}{Dataset} &  \multicolumn{5}{c}{COCA-Net} & \multicolumn{7}{c}{ViT-22B Stack}\\
    \cmidrule(r){1-1}   \cmidrule(r){2-6} \cmidrule(r){7-13} 
    \multicolumn{1}{c}{Name} & \multicolumn{1}{c}{$L$}   & \multicolumn{1}{c}{$d_l$} & \multicolumn{1}{c}{$h_l,w_l$} & \multicolumn{1}{c}{$\tau_l$} & \multicolumn{1}{c}{$k_l$} & \multicolumn{1}{c}{\# of Layers} &  \multicolumn{1}{c}{Q Kernel} &  \multicolumn{1}{c}{Q Stride}   & \multicolumn{1}{c}{Q Pad}  &   \multicolumn{1}{c}{K Kernel} &  \multicolumn{1}{c}{K Stride}   & \multicolumn{1}{c}{K Pad} \\
    \midrule
    
    \multicolumn{1}{c}{Tetrominoes} & \multicolumn{1}{c}{$2$}   & \multicolumn{1}{c}{$[64,64]$} & \multicolumn{1}{c}{$[[4,4],[8,8]]$} & \multicolumn{1}{c}{$[1.00,2.00]$} & \multicolumn{1}{c}{$[3,4]$} & \multicolumn{1}{c}{$[2,2]$} &  \multicolumn{1}{c}{$[4,8]$} &  \multicolumn{1}{c}{$[4,1]$}   & \multicolumn{1}{c}{$[0,0]$}  &   \multicolumn{1}{c}{$[4,8]$} &  \multicolumn{1}{c}{$[4,1]$}   & \multicolumn{1}{c}{$[0,0]$} \\

    \multicolumn{1}{c}{Multi-dSprites} & \multicolumn{1}{c}{$2$}   & \multicolumn{1}{c}{$[64,64]$} & \multicolumn{1}{c}{$[[8,8],[8,8]]$} & \multicolumn{1}{c}{$[1.00,1.25]$} & \multicolumn{1}{c}{$[4,6]$} & \multicolumn{1}{c}{$[3,3]$} &  \multicolumn{1}{c}{$[8,8]$} &  \multicolumn{1}{c}{$[8,1]$}   & \multicolumn{1}{c}{$[0,0]$}  &   \multicolumn{1}{c}{$[8,8]$} &  \multicolumn{1}{c}{$[8,1]$}   & \multicolumn{1}{c}{$[0,0]$} \\

    \multicolumn{1}{c}{ShapeStacks} & \multicolumn{1}{c}{$2$}   & \multicolumn{1}{c}{$[64,64]$} & \multicolumn{1}{c}{$[[8,8],[8,8]]$} & \multicolumn{1}{c}{$[0.75,1.00]$} & \multicolumn{1}{c}{$[4,7]$} & \multicolumn{1}{c}{$[3,3]$} &  \multicolumn{1}{c}{$[8,8]$} &  \multicolumn{1}{c}{$[8,1]$}   & \multicolumn{1}{c}{$[0,0]$}  &   \multicolumn{1}{c}{$[8,8]$} &  \multicolumn{1}{c}{$[8,1]$}   & \multicolumn{1}{c}{$[0,0]$} \\

    \multicolumn{1}{c}{ObjectsRoom} & \multicolumn{1}{c}{$2$}   & \multicolumn{1}{c}{$[64,64]$} & \multicolumn{1}{c}{$[[8,8],[8,8]]$} & \multicolumn{1}{c}{$[1.00,1.50]$} & \multicolumn{1}{c}{$[4,7]$} & \multicolumn{1}{c}{$[3,3]$} &  \multicolumn{1}{c}{$[8,8]$} &  \multicolumn{1}{c}{$[8,1]$}   & \multicolumn{1}{c}{$[0,0]$}  &   \multicolumn{1}{c}{$[8,8]$} &  \multicolumn{1}{c}{$[8,1]$}   & \multicolumn{1}{c}{$[0,0]$} \\

    \multicolumn{1}{c}{CLEVR6} & \multicolumn{1}{c}{$3$}   & \multicolumn{1}{c}{$[96,96,96]$} & \multicolumn{1}{c}{$[[4,4],[4,4],[8,8]]$} & \multicolumn{1}{c}{$[1.0, 0.5, 1.0]$} & \multicolumn{1}{c}{$[2,4,7]$} & \multicolumn{1}{c}{$[1,2,3]$} &  \multicolumn{1}{c}{$[4,4,8]$} &  \multicolumn{1}{c}{$[4,4,1]$}   & \multicolumn{1}{c}{$[0,0,0]$}  &   \multicolumn{1}{c}{$[4,4,8]$} &  \multicolumn{1}{c}{$[4,4,1]$}   & \multicolumn{1}{c}{$[0,0,0]$} \\

     \multicolumn{1}{c}{CLEVRTex} & \multicolumn{1}{c}{$3$}   & \multicolumn{1}{c}{$[96,96,96]$} & \multicolumn{1}{c}{$[[4,4],[4,4],[8,8]]$} & \multicolumn{1}{c}{$[2.0, 1.0, 0.5]$} & \multicolumn{1}{c}{$[2,4,11]$} & \multicolumn{1}{c}{$[1,2,5]$} &  \multicolumn{1}{c}{$[1,1,1]$} &  \multicolumn{1}{c}{$[1,1,1]$}   & \multicolumn{1}{c}{$[0,0,0]$}  &   \multicolumn{1}{c}{$[3,5,8]$} &  \multicolumn{1}{c}{$[1,1,1]$}   & \multicolumn{1}{c}{$[1,2,0]$} \\
    
\bottomrule
  \end{tabular}}
\end{table*}

\subsubsection{Pool, Aggregate and Skip-Connect (Section \ref{sec:pool})}
\paragraph{Pool and Aggregate}
The output cluster masks generated at layer $l$, $\mathbf{\Pi_\mathit{l}} \in \mathbb{R}^{k_l\times n_l}$, provide the COCA layer with the necessary assignments between input nodes and output clusters. Hence, each COCA layer utilizes these output cluster masks to pool attributes from input nodes and construct output cluster attributes. Specifically, we use $\mathbf{\Pi_\mathit{l}}$ to pool a feature vector and five physical attributes per output cluster as:
\begin{subequations}
\begin{align}
\hat{\mathbf{X}}^{l}_{i} &= \frac{\mathbf{\Pi}^{l}_i \mathbf{X}^l + (\mathbf{\Pi}^{l}_i \hat{\mathbf{V}}^l)\mathbf{O}^l + \mathbf{\Pi}^{l}_i \hat{\mathbf{\Psi}}^l} 
{\sum_{j} \mathrm{\Pi}^{l}_{ij} } \label{eqn:supp_agg_line-1}\\
\mathrm{I}^{l+1}_i &= \mathbf{\Pi}^{l}_i\mathbf{I}^{l}\label{eqn:supp_agg_line-2} \\
\mathrm{A}^{l+1}_i &= \mathbf{\Pi}^{l}_i\mathbf{A}^{l}\label{eqn:supp_agg_line-3}  \\
\mathrm{M}^{l+1}_i &= \mathbf{\Pi}^{l}_i\mathbf{M}^{l}\label{eqn:supp_agg_line-4} \\
\mathrm{D}^{l+1}_i &= \mathrm{M}^{l+1}_i / \mathrm{A}^{l+1}_i \label{eqn:supp_agg_line-5}\\
\mathbf{P}^{l+1}_i &= \frac{\mathbf{\Pi}^{l}_i\mathbf{P}^{l}}
{\sum_{j} \mathrm{\Pi}^{l}_{ij} }\label{eqn:supp_agg_line-6}
\end{align}
\end{subequations}
where $\mathbf{O}^l \in \mathbb{R}^{d_l \times d_l}$ represents the output projection, as illustrated in Figure \ref{fig:short-b}, similar to the one used in ViT-22B layer. Note that  $\hat{\mathbf{X}}^{l}_{i}$ still contains the layer index $l$ since final features of the output clusters will be produced after the inter-layer skip connection operation that will be detailed next.

\paragraph{Inter-layer Skip Connections}
To enhance unsupervised hierarchical feature learning, we leverage inter-layer skip connections starting from the second layer and applied at each layer until the hierarchy is completed. Specifically, in order to incorporate the information from node features at layer $l-2$ into cluster features at layer $l$, we first merge the cluster masks that are generated between these two layers, namely masks at layer $l$; $\mathbf{\Pi}^{l} \in \mathbb{R}^{t^2_l \times k_l \times n_l}$ and masks at layer $l-1$; $\mathbf{\Pi}^{l-1} \in \mathbb{R}^{t^2_{l-1} \times k_{l-1} \times n_{l-1}}$. The masks from the preceding layer, $l-1$, are initially un-flattened, reshaped, and partitioned into non-overlapping windows, with each window sized to match the configuration of the current layer $l$. Simultaneously, the masks from the current layer $l$, undergo a process of un-flattening and reshaping to align appropriately. These operations yield the intermediate masks, $\tilde{\mathbf{\Pi}}^{l} \in \mathbb{R}^{t^2_{l} \times (h_{l}w_l) \times k_{l} \times k_{l-1}}$ and $\tilde{\mathbf{\Pi}}^{l-1} \in \mathbb{R}^{t^2_{l} \times (h_{l}w_l) \times k_{l-1} \times (h_{l-1}w_{l-1}k_{l-2})}$, ready for merging:
\begin{equation}
\tilde{\mathbf{\Pi}}^{(l-1)\rightarrow l} = \tilde{\mathbf{\Pi}}^{l} \tilde{\mathbf{\Pi}}^{l-1}
\end{equation}
where $\tilde{\mathbf{\Pi}}^{(l-1)\rightarrow l}$ can be represented as $\tilde{\mathbf{\Pi}}^{(l-1)\rightarrow l} \in \mathbb{R}^{t^2_{l} \times k_{l} \times (h_{l}w_ln_{l-1})}$ after few reshape and transpose operations.

To complete the inter-layer skip connections from layer $l-2$ to $l$, we first skip connect the output cluster features from layer $l-2$; $\hat{\mathbf{X}}^{l-2} \in \mathbb{R}^{t^2_{l-2} \times k_{l-2} \times d_{l-2}}$ and apply appropriate partitioning-- two times to match layer $l$-- and reshaping operations to arrive at $\tilde{\mathbf{X}}^{l-2} \in \mathbb{R}^{t^2_{l} \times (h_{l}w_ln_{l-1}) \times d_{l-2}}$. Note that we keep the feature dimensions of nodes consistent throughout the hierarchy, similar to ViT-22B and shared in Section \ref{sec:suppl_training}, thus we have $d_{l-2}=d_l$. In addition, note that the output cluster features from layer $l=0$ is simply input pixel features generated by the backbone and $h_0=1, w_0=1$.
Having generated the necessary merged masks and reshaped node features, we compute the skip connection from layer $l-2$ to layer $l$ as:
\begin{equation}
\tilde{\mathbf{X}}^{(l-2)\rightarrow l} = \frac{\tilde{\mathbf{\Pi}}^{(l-1)\rightarrow l}\; \tilde{\mathbf{X}}^{l-2}}{\sum_{j} \mathrm{\Pi}^{(l-1)\rightarrow l}_{ij} }
\end{equation}
where $\tilde{\mathbf{X}}^{(l-2)\rightarrow l} \in \mathbb{R}^{t^2_{l} \times k_{l} \times d_l}$. Finally, we apply GroupNorm normalization to $\tilde{\mathbf{X}}^{(l-2)\rightarrow l} $, pass it through an FFN and sum it with the cluster features obtained at layer $l$ (from Eq. \ref{eqn:supp_agg_line-1}), hence complete cluster feature aggregation from layer $l$ to $l+1$.
%\subsection{COCA-Net}
\subsection{Dendrogram Generation by Merging Cluster Masks}
To construct our dendrogram, which represents the object masks eventually generated by the encoder sub-network, we resort to mask merging, similar to the one described in the previous section. Here we progressively merge cluster masks obtained at each level $l$ of the hierarchy, $l=1,...,L$. Specifically, in order to merge the cluster masks that are generated during layer $l$; $\mathbf{\Pi}^{l} \in \mathbb{R}^{t^2_l \times k_l \times n_l}$ and merged masks up until layer $l-1$; $\mathbf{\Pi}^{1\rightarrow (l-1)} \in \mathbb{R}^{t^2_{l-1} \times k_{l-1} \times (\prod_{j=1}^{l-1} h_j w_j)}$, the merged cluster masks from the earlier layer $l-1$ is first un-flattened, reshaped and then divided into non-overlapping windows, with a window size that matches the one at layer $l$. Meanwhile, the masks from the current layer $l$ is also un-flattened and reshaped in an appropriate way. These operations yield the reshaped output cluster mask from layer $l$ as $\tilde{\mathbf{\Pi}}^{l} \in \mathbb{R}^{t^2_{l} \times (h_{l}w_l) \times k_{l} \times k_{l-1}}$ whereas the reshaped merged masks in layer $l-1$ become $\tilde{\mathbf{\Pi}}^{1\rightarrow (l-1)} \in \mathbb{R}^{t^2_{l} \times (h_{l}  w_l) \times k_{l-1} \times (\prod_{j=1}^{l-1} h_j w_j)}$. Now both masks contain the appropriate dimensions to carry out the merging operation as:
\begin{equation}
\tilde{\mathbf{\Pi}}^{1\rightarrow l} = \tilde{\mathbf{\Pi}}^{l} \tilde{\mathbf{\Pi}}^{1\rightarrow (l-1)}
\end{equation}
where $\tilde{\mathbf{\Pi}}^{1\rightarrow l} \in \mathbb{R}^{t^2_{l} \times (h_{l}w_l) \times k_{l} \times (\prod_{j=1}^{l-1} h_j w_j)}$. After appropriate post processing, this merged cluster masks has the dimensions: $\mathbf{\Pi}^{1 \rightarrow l} \in \mathbb{R}^{t^2_{l} \times k_{l} \times (\prod_{j=1}^{l} h_j w_j)}$.

\begin{figure*}[t]
  \vspace{-5pt}
  \centering
  \fbox{\includegraphics[width=.8\textwidth,height=.35\textwidth]{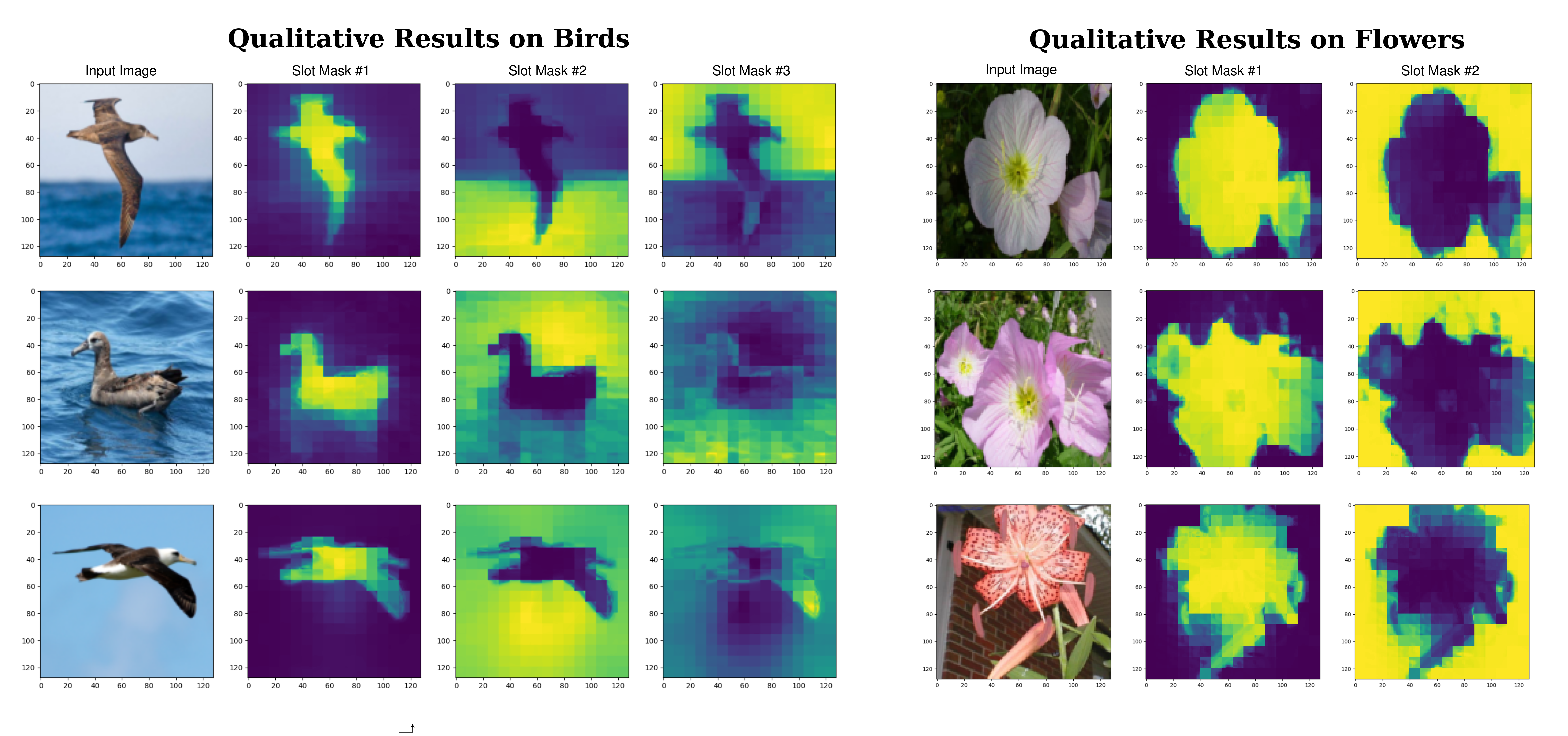}}
  \vspace{-5pt}
   \footnotesize{\caption{Qualitative results obtained for real-world datasets. Slot masks predicted by COCA-Net Encoder are shared for Birds (left) and Flowers (right).}\label{fig:birds_flowers}}
   \vspace{-5pt}
\end{figure*}

\vspace{-5pt}\section{Implementation Details}
\label{sec:suppl_training}
For all experiments included in this work, we build on the comprehensive OCL library that is provided by \cite{Dittadietal22}. This OCL library includes the six datasets that we share results on, in addition to code scripts for training, testing and evaluation. Details are included in \cite{Dittadietal22}. We extend this library by adding the implementations of our proposed architecture COCA-Net and three state-of-the-art methods, which are also our baselines: GEN-v2 \cite{GENESISV2_Engelcke2021}, INVSA \cite{invsa} and BOQSA \cite{boqsa}. To add these baselines to our library, we rely on the official implementation of each baseline and adapt these implementations to work within the OCL library. For a fair comparison, we use a Spatial Broadcast Decoder version for each method.

We train all methods, COCA-Net and the baselines, three times using three random seeds. These three seeds are once generated at random before all experiments and then fixed for all the  methods. Following the training and evaluation criteria laid out in \cite{Dittadietal22}, we train all methods on the same training and validation splits of each dataset, reserving 2000 test images per dataset for evaluation. In each dataset, the maximum number of objects and background segments is used as the number of output slots generated for all methods.

\subsection{Training the COCA-Net}

To optimize the image reconstruction objective on all six datasets, COCA-Net uses the Adam optimizer, employs learning rate linear warm-up and exponential decay scheduling, just as described in the original SA paper \cite{sa}. We set the learning rate as $0.0003$ and adopt weight decay regularization with the coefficient $0.00001$. We train COCA-Net for 500K iterations for each dataset, similar to SA.

For all experiments of the COCA-Net, we use data augmentation as to ``pad and random crop" input images. This augmentation first pads the image with a small number of repeated pixels (e.g., three) on each edge and randomly crops this padded image with a window that has the same resolution as the original image. The motivation behind this particular augmentation strategy is to present each COCA layer with a randomly shifted non-overlapping partitions, effectively increasing the window variability that COCA-Net learns from.

Tables \ref{tbl:hyperparameter_backbone} and \ref{tbl:hyperparameter_sbd} list the hyper-parameters used across all six datasets for the Pixel Feature Encoder Backbone and the Spatial Broadcast Decoder, respectively. Table \ref{tbl:hyperparameter_cocanet} presents the hyper-parameters employed for COCA-Net on each dataset. 

\subsection{Training the Baseline Models}
For all baselines, we use the default configurations and hyper-parameters that are laid out in their original papers. Unlike some of our baselines, we train all methods to produce output object masks that match the ground-truth maximum number of objects in a dataset. In addition, due to computational constraints, we fix the batch size as 64 for all methods.

\section{Additional Results}
\label{sec:suppl_add_results}
\subsection{Initial Experiments on Real-World Datasets; Birds and Flowers}
\label{sec:suppl_init_real}
We evaluated COCA-Net on Birds \cite{Birds} and Flowers \cite{Flowers} (following BOQSA's protocol) using the SLATE \cite{Dalle} architecture which leverages a transformer decoder but no pretrained backbone. Entire architecture is trained from scratch. Both datasets contain real-world images with compositionality (objects contain a spectrum of colors and deformable shapes). Table \ref{tbl:suppl_quant_real} shows COCA-Net is on par or better than BOQSA across IoU and Dice metrics. Figure \ref{fig:birds_flowers} displays qualitative results, with COCA-Net achieving robust segmentation masks for both foreground objects and background segments. Since these are our initial experiments with COCA-Net integrated into the SLATE architecture, we believe that the performance can be further boosted with more elaborate hyperparameter tuning.

\begin{table*}
  \centering
  \hspace*{-2.em}
  \vspace{-10pt}
  \footnotesize{\caption{\footnotesize{Quantitative Results for COCA-Net and BOQSA on Birds and Flowers datasets. Background segments included, single seed set to BOQSA's open-source implementation.}} \label{tbl:suppl_quant_real}}
  \begin{tabular}{lllll}
    \toprule
    \multicolumn{1}{c}{} & \multicolumn{2}{c}{Flowers} & \multicolumn{2}{c}{Birds} \\
    \midrule 
    \multicolumn{1}{c}{Model} & \multicolumn{1}{c}{IoU$\uparrow$} & \multicolumn{1}{c}{Dice$\uparrow$} & \multicolumn{1}{c}{IoU$\uparrow$} & \multicolumn{1}{c}{Dice$\uparrow$}\\
    \midrule 
    BOQSA &  \multicolumn{1}{c}{0.6477} & \multicolumn{1}{c}{0.7603} &  \multicolumn{1}{c}{0.5565} & \multicolumn{1}{c}{0.7032}\\    
    COCA-Net (ours) &  \multicolumn{1}{c}{\textbf{0.6482}} & \multicolumn{1}{c}{\textbf{0.7612}} &  \multicolumn{1}{c}{\textbf{0.5778}} & \multicolumn{1}{c}{\textbf{0.7238}}\\
    \bottomrule
  \end{tabular}
\vspace{-10pt}
\end{table*}

\subsection{Additional Quantitative Results}
\label{sec:suppl_quant}
Here, we present Table \ref{tbl:supp_quant},  an extension of Table \ref{tbl:tetro} that was introduced in the main paper. Table \ref{tbl:supp_quant} provides additional quantitative results of COCA-Net and its baselines obtained on the Tetrominoes and Multi-dSprites datasets.
\begin{table*}[h]
  \footnotesize{
  \caption{\footnotesize{Supplementary unsupervised scene segmentation results of three baseline models and proposed COCA-Net on Tetrominoes and Multi-dSprites datasets, based on nine performance evaluation metrics, assessed across four different configurations. Scores are reported as mean $\pm$ standard deviation for 3 seeds.}}
  \centering
  \hspace*{1.em}
  \label{tbl:supp_quant}
  \begin{tabular}{l l l l l l l l l l }
    \toprule
    & \multicolumn{2}{c}{DEC-FG Only} & \multicolumn{2}{c}{DEC-BG Included} & \multicolumn{2}{c}{ENC-FG Only} & \multicolumn{2}{c}{ENC-BG Included} \\
    \cmidrule(r){2-3}  \cmidrule(r){4-5}  \cmidrule(r){6-7}  \cmidrule(r){8-9}
    \multicolumn{1}{c}{Name} & \multicolumn{1}{c}{ARI$\uparrow$} & \multicolumn{1}{c}{mSC$\uparrow$}   &  \multicolumn{1}{c}{ARI$\uparrow$} &  \multicolumn{1}{c}{mSC$\uparrow$} &  \multicolumn{1}{c}{ARI$\uparrow$}   & \multicolumn{1}{c}{mSC$\uparrow$}  & \multicolumn{1}{c}{ARI$\uparrow$}   & \multicolumn{1}{c}{mSC$\uparrow$}      &  \multicolumn{1}{c}{MSE$\downarrow$} \\
    \midrule
    \multicolumn{10}{c}{Tetrominoes} \\
    \midrule
    GEN-v2 \cite{GENESISV2_Engelcke2021}  & 0.33$\pm$0.47	 &      0.17$\pm$0.10	 & 0.03$\pm$0.04	 & 	0.27$\pm$0.03	 & 0.13$\pm$0.18		 & 	0.13$\pm$0.05	 & 0.02$\pm$0.02		 & 0.24$\pm$0.02	 & 0.004$\pm$0.000\\
    INV-SA \cite{invsa} & 0.98$\pm$0.01 &		0.97$\pm$0.00 &	0.97$\pm$0.00 &		0.98$\pm$0.00 &	0.38$\pm$0.08 &		0.59$\pm$0.05 &	0.61$\pm$0.04 &		0.65$\pm$0.04 &	0.001$\pm$0.000 \\
    BOQ-SA \cite{boqsa} & \textbf{0.99$\pm$0.01}	& 	0.31$\pm$0.00	& 0.11$\pm$0.01	& 	0.33$\pm$0.01	& 0.60$\pm$0.03		& 0.42$\pm$0.01	& 0.29$\pm$0.02		& 0.47$\pm$0.00	& \textbf{0.000$\pm$0.000}\\
    COCA-Net (ours)   & \textbf{0.99$\pm$0.01} &		\textbf{0.98$\pm$0.01} &	\textbf{0.99$\pm$0.01} &		\textbf{0.99$\pm$0.01} &	\textbf{0.74$\pm$0.04} &		\textbf{0.70$\pm$0.07} &	\textbf{0.71$\pm$0.09} &		\textbf{0.75$\pm$0.07} &	0.001$\pm$0.000  \\
    \midrule
    \multicolumn{10}{c}{Multi-dSprites} \\
    \midrule
    GEN-v2 \cite{GENESISV2_Engelcke2021} & 0.80$\pm$0.03	&	0.58$\pm$0.04	& 0.71$\pm$0.02	&	0.66$\pm$0.03 & 0.60$\pm$0.02	&	0.15$\pm$0.02 &	0.04$\pm$0.01	&	0.22$\pm$0.01 &	0.007$\pm$0.000 \\
    INV-SA\cite{invsa} & 0.90$\pm$0.00 &		0.84$\pm$0.01 &	0.71$\pm$0.35 &		0.84$\pm$0.05 &	0.68$\pm$0.04 &		0.47$\pm$0.00 &	0.41$\pm$0.26 &		0.53$\pm$0.06 &	\textbf{0.001$\pm$0.000} \\
    BOQ-SA \cite{boqsa} & 0.89$\pm$0.00 &		0.65$\pm$0.07 &	0.42$\pm$0.17 &		0.65$\pm$0.10 &	0.75$\pm$0.01 &		0.55$\pm$0.01 &	0.34$\pm$0.06 &		0.56$\pm$0.02 &	\textbf{0.001$\pm$0.000} \\
    COCA-Net (ours) & \textbf{0.95$\pm$0.00}&		\textbf{0.91$\pm$0.01} &	\textbf{0.84$\pm$0.19} &	\textbf{0.91$\pm$0.03} &	\textbf{0.96$\pm$0.01} &		\textbf{0.95$\pm$0.01} &	\textbf{0.98$\pm$0.00}&		\textbf{0.96$\pm$0.00} &	0.002$\pm$0.000 \\   
\bottomrule
  \end{tabular}}
\end{table*}

\begin{table*}[h]
  \centering
  \footnotesize{
  \caption{\footnotesize{Extension of Table \ref{tbl:ablation} from the main paper, comparing the unsupervised scene segmentation performance of COCA-Net and COCA-Net-RAS on two datasets. The comparison is based on nine performance evaluation metrics assessed across four different configurations, with scores reported for the same single seed.}}
  \hspace*{1.em}
  \label{tbl:supp_abl_ras}
  \begin{tabular}{l l l l l l l l l l }
    \toprule
    & \multicolumn{2}{c}{DEC-FG Only} & \multicolumn{2}{c}{DEC-BG Included} & \multicolumn{2}{c}{ENC-FG Only} & \multicolumn{2}{c}{ENC-BG Included} \\
    \cmidrule(r){2-3}  \cmidrule(r){4-5}  \cmidrule(r){6-7}  \cmidrule(r){8-9}
    \multicolumn{1}{c}{Name} & \multicolumn{1}{c}{ARI$\uparrow$} & \multicolumn{1}{c}{mSC$\uparrow$}   &  \multicolumn{1}{c}{ARI$\uparrow$} &  \multicolumn{1}{c}{mSC$\uparrow$} &  \multicolumn{1}{c}{ARI$\uparrow$}   & \multicolumn{1}{c}{mSC$\uparrow$}  & \multicolumn{1}{c}{ARI$\uparrow$}   & \multicolumn{1}{c}{mSC$\uparrow$}      &  \multicolumn{1}{c}{MSE$\downarrow$} \\
    \midrule
    \multicolumn{10}{c}{ObjectsRoom} \\
    \midrule
    COCA-Net-RAS & 0.832	& 0.739	& 	0.561	& 0.581	& 	0.763	& 0.277		& 0.469	& 0.368		& \textbf{0.001} \\
    COCA-Net    & \textbf{0.894}	& \textbf{0.832}		& \textbf{0.960}	& \textbf{0.889}		& \textbf{0.879}	& \textbf{0.823}		& \textbf{0.952}	& \textbf{0.881}	& 	\textbf{0.001} \\
    \midrule
    \multicolumn{10}{c}{ShapeStacks} \\
    \midrule
    COCA-Net-RAS & 0.834	& 0.735		& \textbf{0.267}	& 0.728		& 0.747	& 0.385		& 0.157	& 0.419		& 0.005 \\
    COCA-Net    & \textbf{0.916}	& \textbf{0.857}	& 	0.230	& \textbf{0.782}		& \textbf{0.843}	& \textbf{0.865}	& 	\textbf{0.209}	& \textbf{0.772}		& \textbf{0.004}\\
\bottomrule
  \end{tabular}}
\end{table*}

\begin{table*}[h]
  \centering
  \footnotesize{
    \caption{\footnotesize{Extension of Table \ref{tbl:ablation} from the main paper, comparing the unsupervised scene segmentation performance of COCA-Net and COCA-Net-Dyna on two datasets. The comparison is based on nine performance evaluation metrics assessed across four different configurations. All scores are based on results obtained using the same random seed. The average number of slots used is provided in the second column. Note that for the CLEVR dataset, COCA-Net is trained and evaluated on CLEVR6, whereas COCA-Net-Dyna is trained on CLEVR6 but evaluated on CLEVR10.}}
  \hspace*{-2.5em}
\label{tbl:supp_abl_dyna}
  \begin{tabular}{l l l l l l l l l l l }
    \toprule
    & &\multicolumn{2}{c}{DEC-FG Only} & \multicolumn{2}{c}{DEC-BG Included} & \multicolumn{2}{c}{ENC-FG Only} & \multicolumn{2}{c}{ENC-BG Included} \\
    \cmidrule(r){3-4}  \cmidrule(r){5-6}  \cmidrule(r){7-8}  \cmidrule(r){9-10}
    \multicolumn{1}{c}{Name} & \multicolumn{1}{c}{Avg. Slots} & \multicolumn{1}{c}{ARI$\uparrow$} & \multicolumn{1}{c}{mSC$\uparrow$}   &  \multicolumn{1}{c}{ARI$\uparrow$} &  \multicolumn{1}{c}{mSC$\uparrow$} &  \multicolumn{1}{c}{ARI$\uparrow$}   & \multicolumn{1}{c}{mSC$\uparrow$}  & \multicolumn{1}{c}{ARI$\uparrow$}   & \multicolumn{1}{c}{mSC$\uparrow$}      &  \multicolumn{1}{c}{MSE$\downarrow$} \\
    \midrule
    \multicolumn{10}{c}{CLEVR} \\
    \midrule
    COCA-Net & \multicolumn{1}{c}{11}  &  0.982	& 0.881& 		0.929& 	0.900	& 	0.975	& 0.756		& 0.849	& 0.795	& 	0.000 \\
    COCA-Net-Dyna & \multicolumn{1}{c}{7.49}  &  0.978	& 0.840	& 	0.911	& 0.859	& 	0.966	& 0.741		& 0.846	& 0.773		& 0.001 \\
    \midrule
    \multicolumn{10}{c}{ShapeStacks} \\
    \midrule
    COCA-Net & \multicolumn{1}{c}{7}  & 0.916	& 0.857		& 0.230	& 0.782		& 0.843	& 0.865		& 0.209	& 0.772 & 0.004 \\
    COCA-Net-Dyna & \multicolumn{1}{c}{5.35} & 0.896	& 0.818		& 0.232	& 0.754		& 0.827	& 0.823		& 0.210	& 0.746 & 0.005 \\
\bottomrule
  \end{tabular}}
\end{table*}

\subsection{Ablation Studies}
\label{sec:suppl_abl}
We now present additional quantitative results to complement those in Table \ref{tbl:ablation} of the main paper. Table \ref{tbl:supp_abl_ras} extends our comparison between COCA-Net and its random anchor node selection variant, COCA-Net-RAS. Further results for COCA-Net and COCA-Net-Dyna are provided in Table \ref{tbl:supp_abl_dyna}.

\subsection{Additional Qualitative Results}
\label{sec:suppl_qual}
\begin{figure*}[t!]
  \centering
  \fbox{\includegraphics[width=.5\textwidth,height=1.2\textwidth]{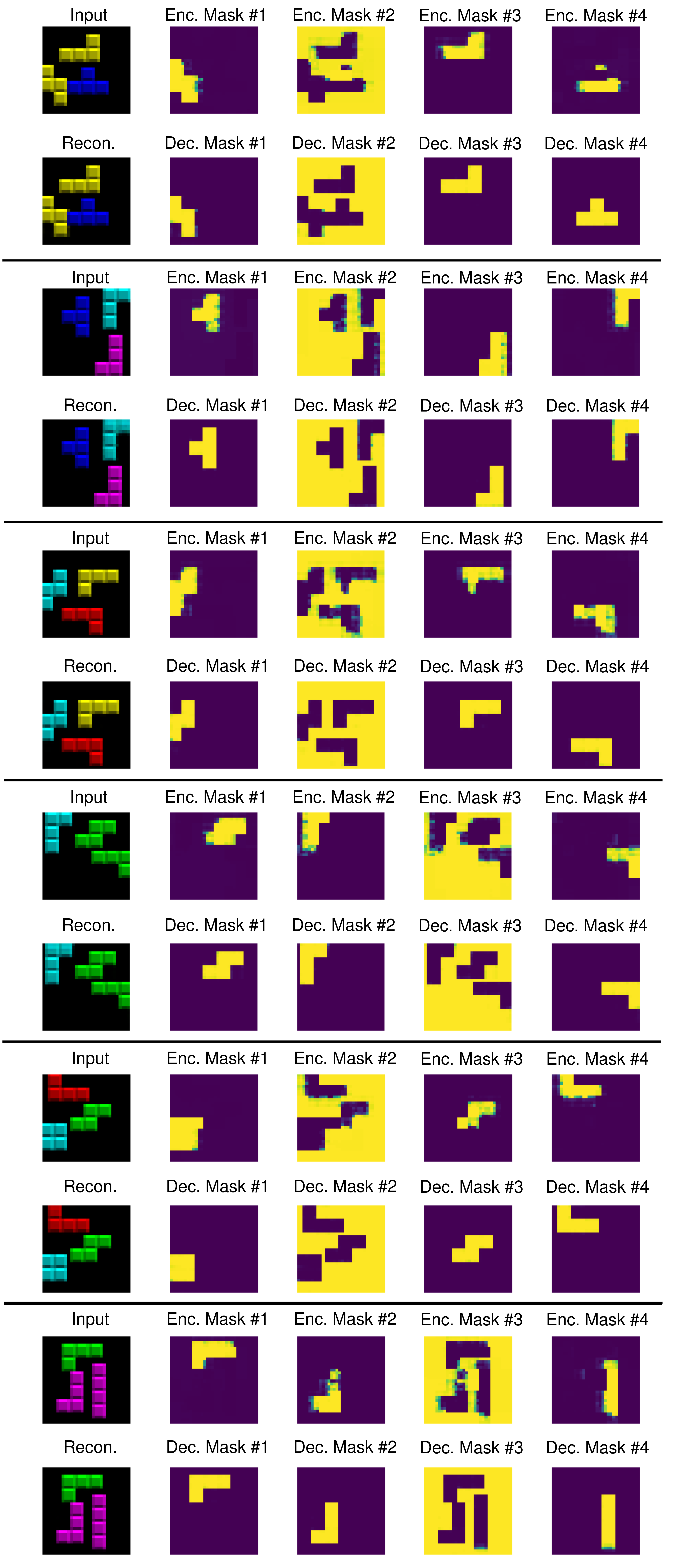}}
    \caption{Qualitative results of COCA-Net on Tetrominoes dataset. Accompanying each input image, we visualize COCA-Net's reconstruction, segmentation masks generated by its encoder sub-network, as well as segmentation masks predicted by the decoder sub-network.}
    \label{fig:qualitative_te}
\end{figure*}

\begin{figure*}[h]
  \centering
  \fbox{\includegraphics[width=.6\textwidth,height=1.25\textwidth]{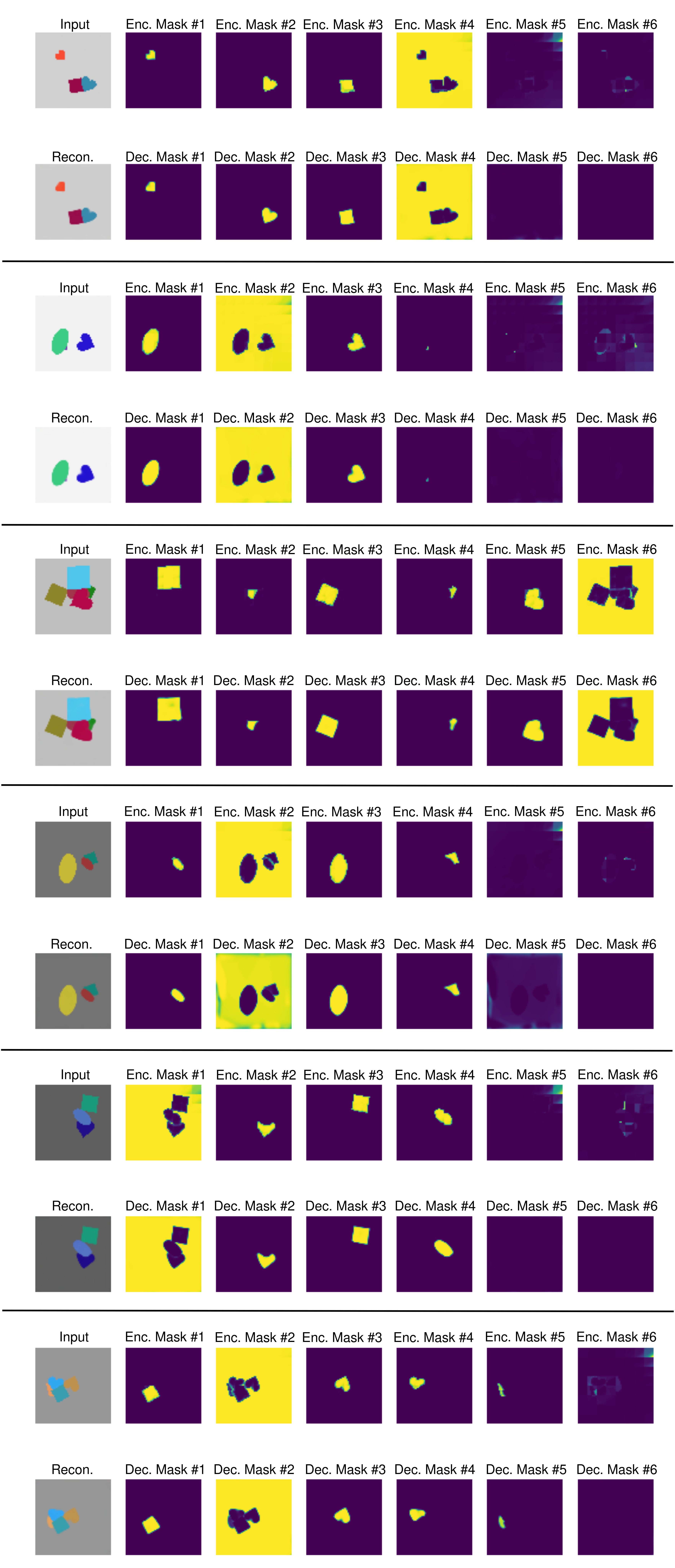}}
    \caption{Qualitative results of COCA-Net on Multi-dSprites dataset. Accompanying each input image, COCA-Net's reconstruction, segmentation masks generated by its encoder sub-network, as well as segmentation masks predicted by the decoder sub-network are visualized.}
    \label{fig:qualitative_ds}
\end{figure*}

\begin{figure*}[h]
  \centering
  \fbox{\includegraphics[width=.75\textwidth,height=1.25\textwidth]{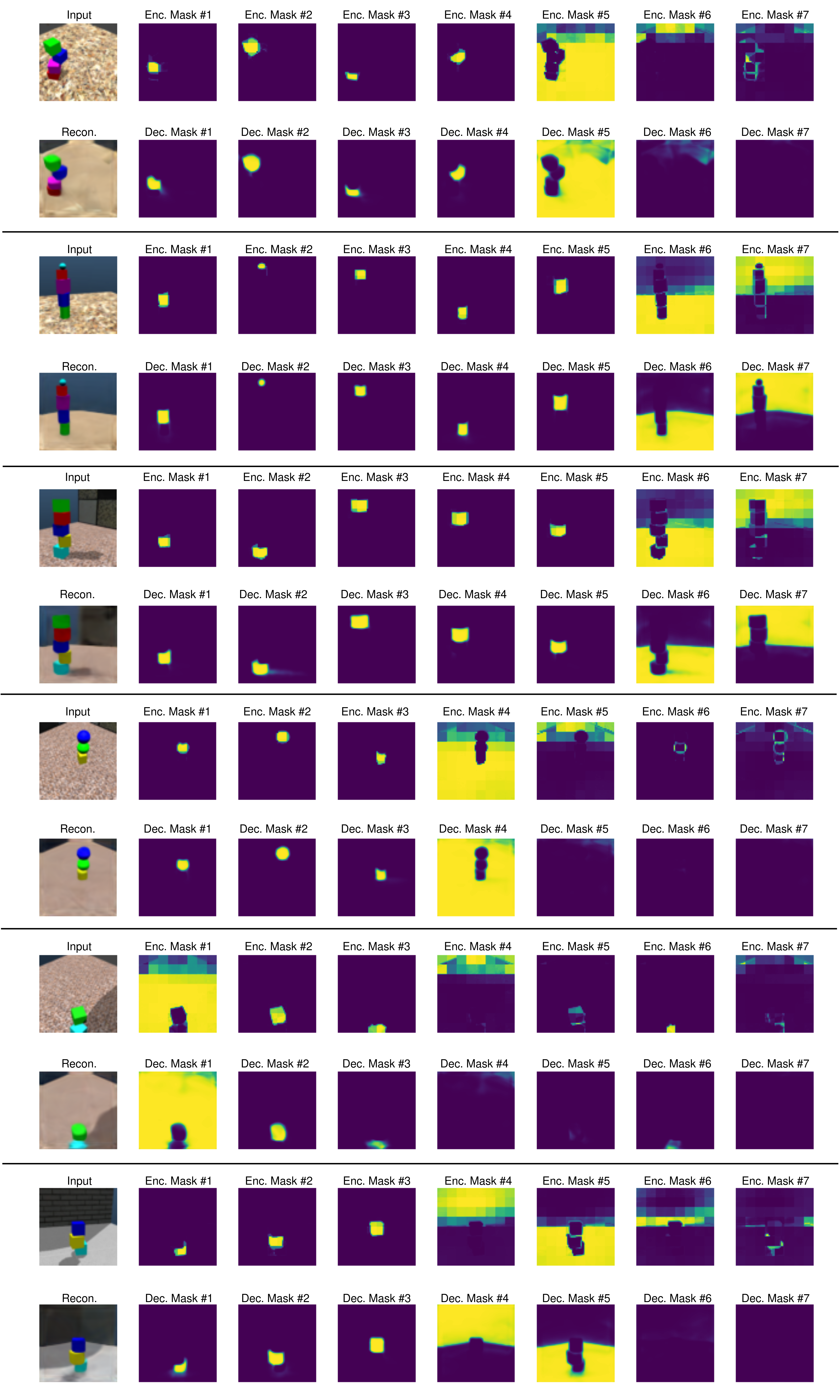}}
    \caption{Qualitative results of COCA-Net on Shapestacks dataset. With each input image, we visualize COCA-Net's reconstruction, segmentation masks generated by its encoder sub-network, as well as segmentation masks predicted by the decoder sub-network.}
    \label{fig:qualitative_ss}
\end{figure*}

\begin{figure*}[h]
  \centering
  \fbox{\includegraphics[width=.75\textwidth,height=1.25\textwidth]{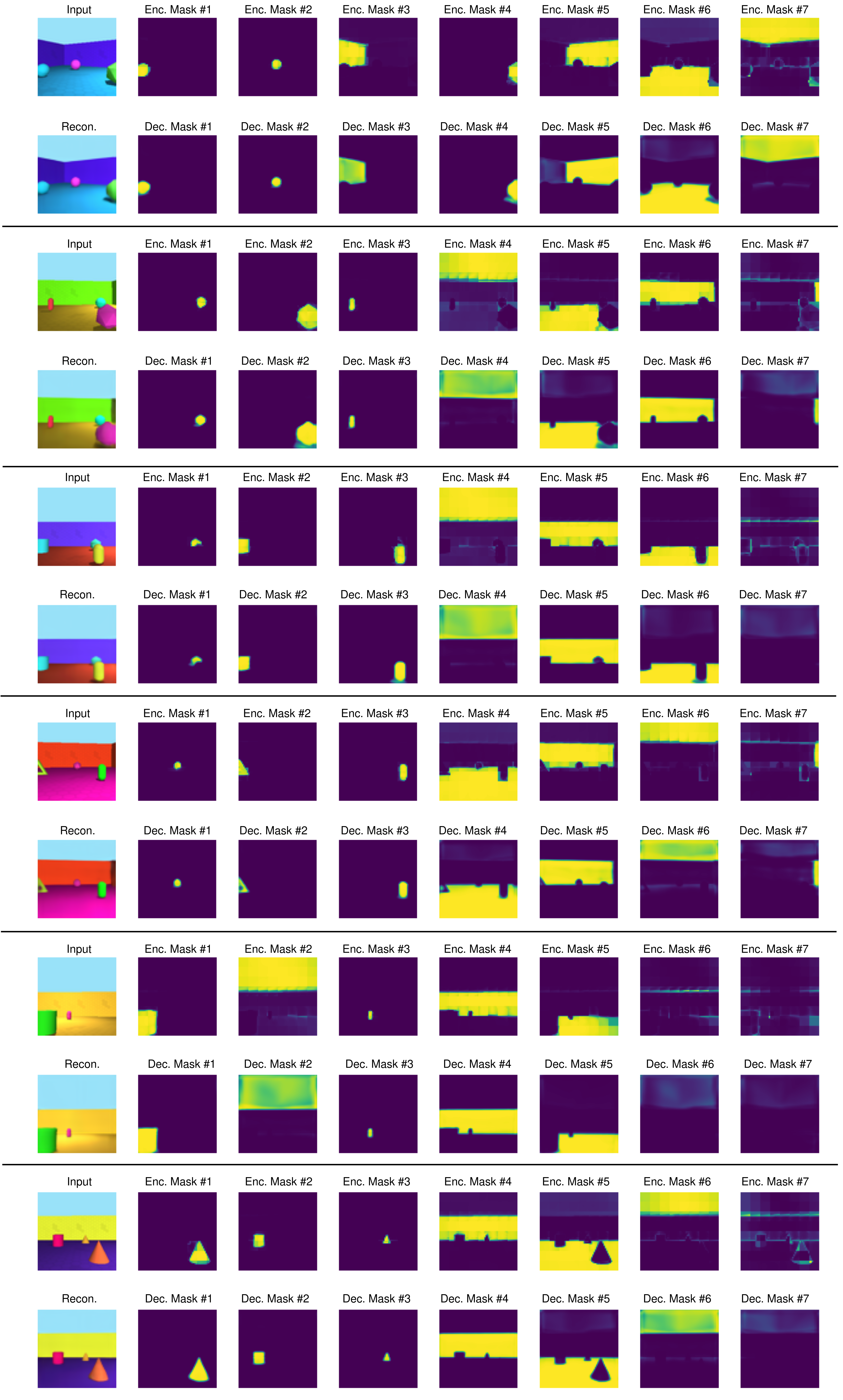}}
    \caption{Qualitative results of COCA-Net on ObjectRoom dataset. With each input image, we visualize COCA-Net's reconstruction, segmentation masks generated by its encoder sub-network, as well as segmentation masks predicted by the decoder sub-network.}
    \label{fig:qualitative_or}
\end{figure*}

\begin{figure*}[h]
  \centering
  \fbox{\includegraphics[width=.7\textwidth,height=1.25\textwidth]{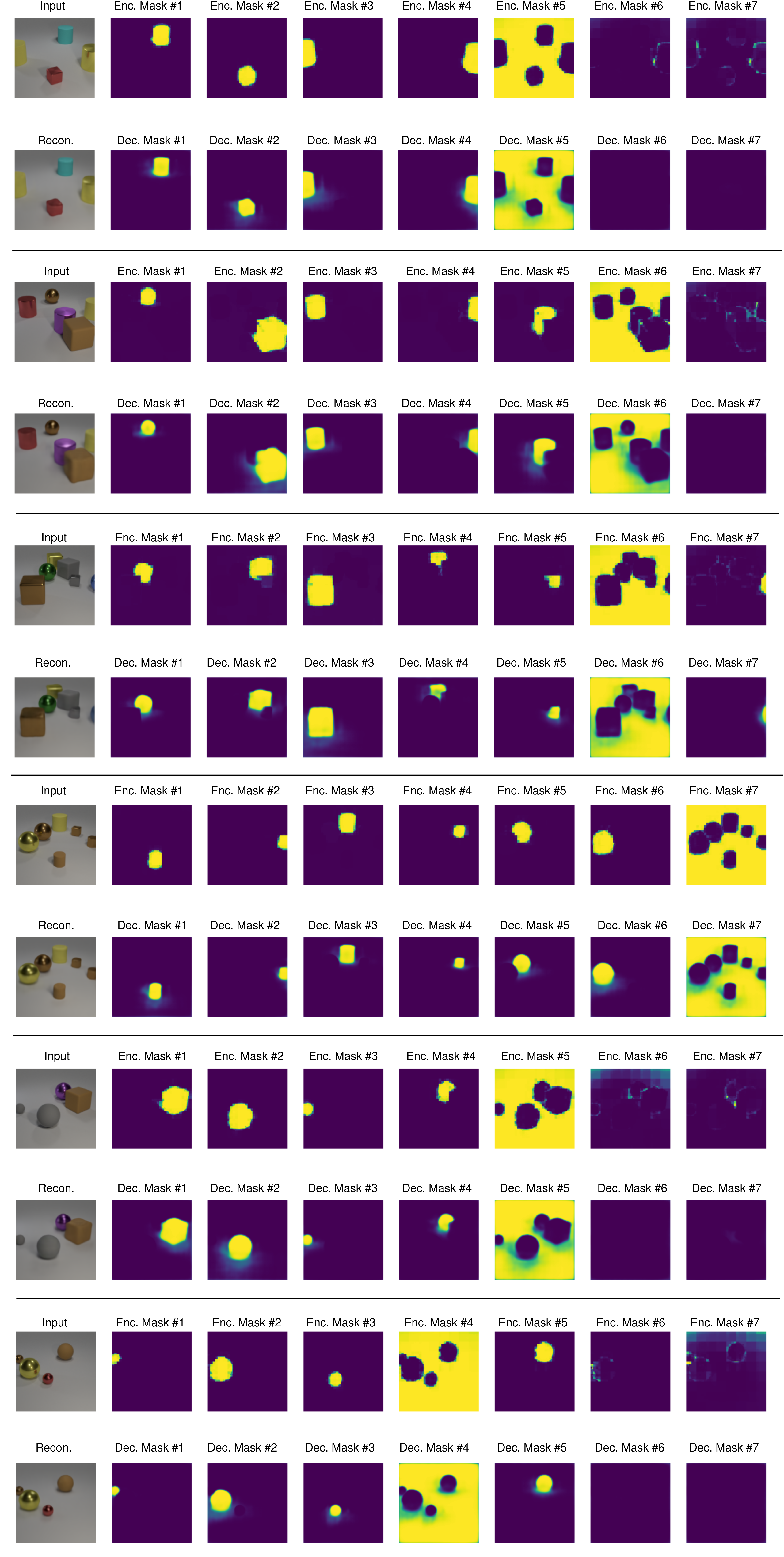}}
    \caption{Qualitative results of COCA-Net on CLEVR6 dataset. With each input image, we visualize COCA-Net's reconstruction, segmentation masks generated by its encoder sub-network, as well as segmentation masks predicted by the decoder sub-network.}
    \label{fig:qualitative_cl}
\end{figure*}

\begin{figure*}[h]
  \centering
   \hspace*{-4.5em}
  \fbox{\includegraphics[width=1.15\textwidth,height=1.25\textwidth]{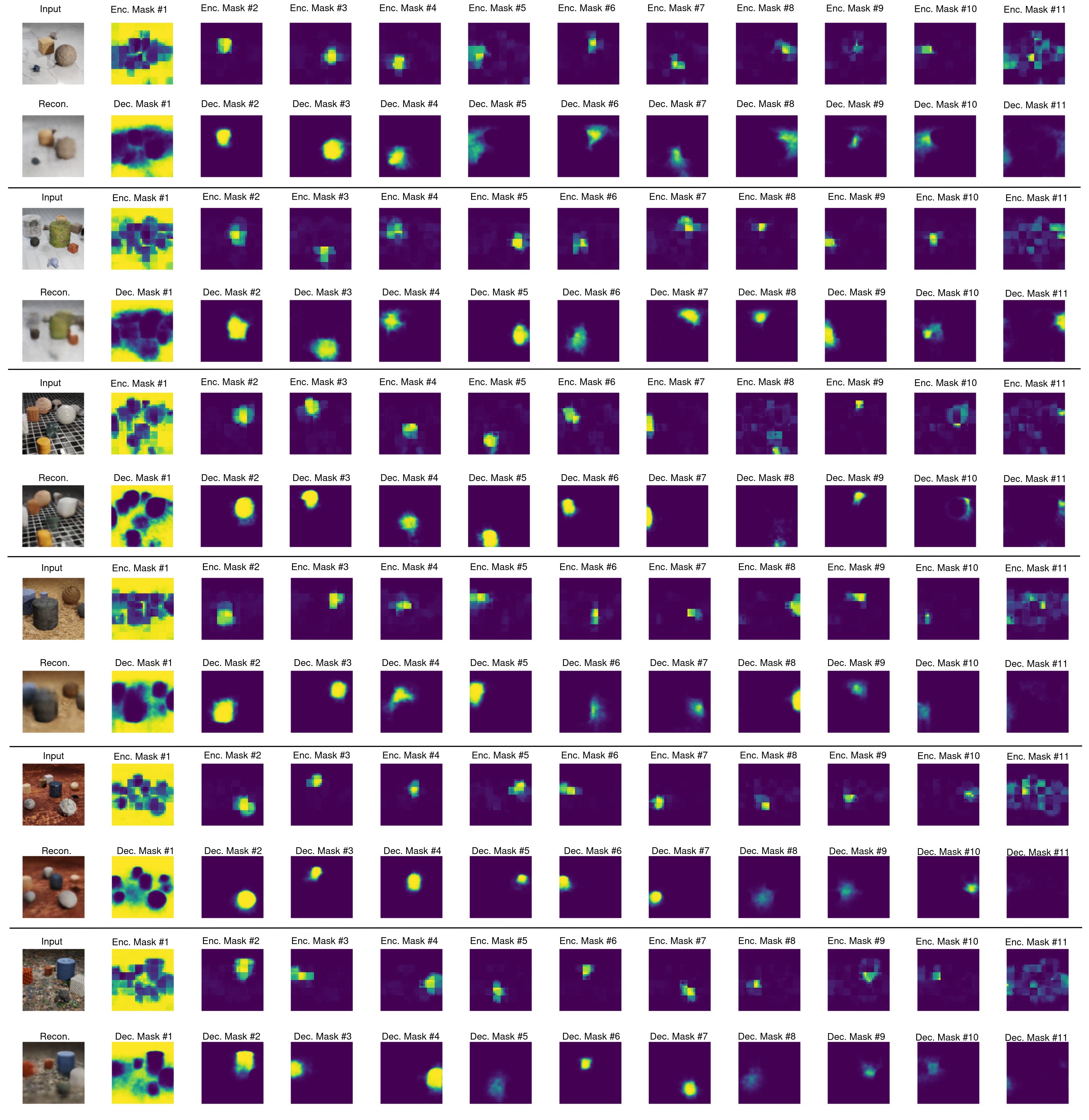}}
    \caption{Qualitative results of COCA-Net on CLEVRTex dataset. With each input image, we visualize COCA-Net's reconstruction, segmentation masks generated by its encoder sub-network, as well as segmentation masks predicted by the decoder sub-network.}
    \label{fig:qualitative_ct}
\end{figure*}

We share the supplementary qualitative results of COCA-Net, starting from Figure \ref{fig:qualitative_te} to Figure \ref{fig:qualitative_ct} for all six datasets. In each figure, we share a sample batch of images from the corresponding dataset and include COCA-Net's reconstructions of these images next to it. In addition, we visualize the segmentation masks that COCA-Net produces in its encoder and decoder sub-networks.


\end{document}